# From PID to Reinforcement Learning in Building HVAC Control: A Control-Architecture Review of Tuning Strategies, Formulations, and Deployment Trade-offs

Erfan Haghighat Damavandi[1,2*], Davide Papurello[2], Mahdi Alibeigi[3], Armin Keshavarz[3]

[1] *R&D Deprtment, Owtana Tech, Torino, Italy*
[2] *Energy Department, Politecnico di Torino, Torino, Italy*
[3] *Energy Department, Uversity Tehran, Tehran, Iran*

*Corresponding author. E-mail address: in@owtantech.com

---

**Abstract**

This paper reviews control architectures for building heating, ventilation and air-conditioning (HVAC) systems, addressing the tension among energy demand, occupant comfort and equipment operability, and situating proportional-integral-derivative (PID) regulation within a layered control stack that also includes rule-based and model-based supervisory control and data-driven reinforcement learning (RL). It examines the thermostat–PID layer from two-position and scheduled setpoint control through proportional, PI and PID regulation and their tuning, then formulates RL for building control as a Markov decision process. Its central contribution is a taxonomy of where RL meets PID: RL as a gain tuner for an existing loop, as a direct replacement for the regulatory controller, as a hybrid with model predictive control, and as supervisory setpoint and schedule generation. Across these classes, the review establishes that the physical loop remains invariant while the intervention point changes, and that the classes differ in their requirements for actuation authority, state information, modelling, commissioning and constraint handling. It further establishes that PID is inexpensive and well codified but limited by nonlinearity, time variation, discrete mode switching, loop interaction and slow plant dynamics, while RL's admissible position must be argued rather than assumed. The paper concludes that these architectures are not competing philosophies of control but nested points of attachment on one HVAC plant, with hybrid implementations possible, and that deployment trade-offs must be assessed according to the specific control interface and scaffolding required.

---

## 1. Introduction

### *1.1 Introduction*

Control of heating, ventilation and air-conditioning (HVAC) plant sits at the intersection of three concerns that building research usually treats separately: energy demand, occupant comfort, and the operability of mechanical equipment. Comfort is not a soft add-on but a measurable outcome of plant behaviour: in a survey of two non-residential Norwegian buildings, the most important factors affecting occupant comfort were poor air quality, lack of natural light and uncomfortable temperature, rated 9, 9 and 8 out of 10, respectively [1]. Operability is bound up with the same decisions, since improper planning of a building's

HVAC system due to a failure to account for its intended use and expected population is a leading cause of HVAC breakdowns [1]. On the energy side, the regulatory layer is where leverage is greatest: indoor temperature control through occupancy-driven thermostats based on adaptive comfort principles can significantly reduce energy consumption, by 30–50 %, even in air-conditioned buildings [2]. Comfort, faults and energy therefore respond to the same variable — the temperature setpoint and the law that tracks it — which is why the architecture of that law, and not only its individual algorithms, merits review.

That architecture is layered, and each layer has its own established practice. At the setpoint layer, the conventional method fixes the indoor temperature at 20 °C in winter and 24 °C in summer according to the ASHRAE 55 standard, whereas in the adaptive setpoint approach the indoor temperature is not fixed but changes proportionately to the monthly average of the outdoor temperature [2]. At the regulatory layer, proportional-integral-derivative (PID) controllers remain the most widely adopted and popular controllers in industry because of their simplicity, robustness, low cost and broad applicability with satisfactory performance, particularly at the lower level of the control hierarchy [3]. At the supervisory layer, the traditional approaches are rule-based control (RBC) and model predictive control (MPC), with machine learning opening up a new route of data-driven control (DDC) [4]. Reinforcement learning (RL) is one such route: an approach that employs algorithms or "agents" to navigate sequential decision making in an environment with limited feedback, and one that is especially suitable for defining policies and behaviours [4]. Read as a stack, these layers are complementary; read as a menu, they appear to compete.

The literature has largely read them as a menu. Positions on machine learning versus physics-based building performance simulation (BPS) are openly adversarial: Geyer and Singravel and Kathirgamanathan et al. voice the belief that machine learning models will replace traditional physics-based BPS models, while Roth et al. counter that "supervised machine learning models (…) are not a silver bullet for energy prediction, despite the recent rapid advancements in these tools. These models are often difficult to interpret and provide little physical explanatory power" [4]. The adversarial framing has a sociological signature: Arroyo et al. observe that authors often "use either one approach or another based on their expertise and justify their choice by highlighting the strengths of their approach and stressing the weaknesses of the other" [4]. Within control specifically, the same authors note that "the control and machine learning communities keep evolving independently with a radically different notation natively adopted to formulate the same problem", in reference to MPC and DDC approaches [4]. The ladder narrative is further complicated by the observation that the new emergent digital domains, while enjoying significant traction, in part offer new terminology for existing technology, machine learning having long been around as "black box" approaches with increased availability of training data as the main difference between past and present [4]. Against this, Deb and Schlueter identify an opportunity to integrate black box, grey box and white box models in a common frame that exploits the benefits of each approach [4].

The problem this review addresses follows directly from that observation: RL is habitually inserted into the sequence as the successor to PID rather than as a component whose admissible position in the stack must be argued. Three positions are, in principle, available. RL can act as a supervisory PID tuner, replacing or augmenting conventional tuning rules — rule-based approaches such as Ziegler-Nichols and Cohen-Coon, and model-based approaches such as Internal Model Control (IMC) — which become less applicable in nonlinear and time-varying plants because they require frequent re-tuning whenever the system changes [3]. The evidence base for this position exists: in a nonlinear multiple-input, multiple-output (MIMO) continuous stirred-tank reactor subject to time-varying uncertainty, conventional PID failed to track the effluent concentration setpoint and caused large errors and offsets, whereas the RL agents achieved fast and accurate setpoint tracking that significantly reduced the errors and eliminated offsets [3]. RL can act at the setpoint and thermostat layer, where adaptive strategies have already been shown to shift envelope requirements — the ASHRAE adaptive setpoint approach reduced the optimal insulation thickness by up to 66 % depending on climate type, local adaptive setpoints eliminated the need for thermal insulation in all Iran climates and all three common wall materials studied, and the purchase, installation and maintenance costs of adaptive thermostats were recoverable in less than two years [2] — and RL is indeed identified as an important approach for the control of buildings, with particular advantages where real-time interaction is required [4]. Or RL can act as a direct control policy, which is the framing implied by its definition as an agent navigating sequential decisions under limited feedback [4].

Each position carries a different scaffolding requirement, and physics-based models are the scaffolding in question. A digital twin becomes useful where a model, as digital counterpart, can be used to manage a real building, for instance for optimal control or predictive maintenance [4]; one implementation integrates building information modelling (BIM) with real-time sensor data, occupant feedback and a probabilistic model of occupant comfort to detect and predict HVAC issues that may impact comfort, using 200 000 points of historical sensor data [1]. The performance claims attached to such scaffolding are concrete: a BIM plugin that handles data every 5 min was previously unavailable, and the updated plugin achieved a 35 % reduction in the time required to identify and resolve HVAC faults compared

## 2. System Description: HVAC Plant and Control-Loop Anatomy

### *2.1 System Description: HVAC Plant and Control-Loop Anatomy*

#### *2.1.1 Reference plant configuration*

Throughout this review the air-side plant of building NE01 at the British Columbia Institute of Technology (Burnaby, BC, Canada) serves as the reference configuration. NE01 is a four-storey educational building of 20,076.88 m² gross floor area, and the temperatures in its zones are controlled by ten independent air-handling units (AHUs) [5]. Of these, AHU7 delivers air to seven variable air volume (VAV) units serving seven interior zones on the fourth floor, comprising six classrooms and an office area, and it is the data and parameters of AHU7 that form the basis of the running example [5]. A building management system (BMS) collects the data from the sensors and virtual meters of the air-handling system [5]. The plant was selected because of its high energy consumption relative to other buildings on the campus, its alignment with institutional sustainability goals, the availability of archived data over many years for multiple zones served by the same system, and the potential to demonstrate energy savings and improved occupant experience [5].

Airflow is controlled through the supply and return fans; recirculated air from the seven service zones is mixed with outdoor air and reconditioned through the AHU, with the mixed-air condition regulated by three dampers that set the percentages of air exhausted from the system, entering the system, and recirculating within it [5]. The mixed air is drawn through the filter by the supply fan and is then cooled or heated as it passes over the cooling and heating coils; the heating coil is supplied with hot water from the boiler, and heating output is regulated by adjusting the water flow through the coil by means of an electronic valve [5]. A chiller and a heater provide chilled and hot water to the VAV systems, the fresh air being circulated through the AHU and leaving through the exhaust system, while the cooling and heating load automatically controls the VAV system [6]. Commercial plants of this type typically combine several layers of air distribution — air-handling units, fans and ducts — with a wide range of water-based systems such as cooling towers, chillers and boilers, and this multiplicity of zones and systems is what makes the development of plant models demanding [7]. The physical distribution of these items is likewise characteristic: the system control panel, boilers, motors, pumps and chillers are often located on the lower level, the cooling tower on the roof, and ductwork, fans, dampers, coils, air filters, heating units, VAV units and diffusers are distributed throughout the building, with larger buildings often having separate systems for groups of floors or areas [8].

For the analyses that follow, the air-side process can be summarised by three relations. Neglecting latent effects and assuming a constant specific heat, the mixed-air temperature is the flow-weighted average of the outdoor and return streams, where $x_{oa}$ is the outdoor-air fraction set by the damper positions:

$$T_{ma} = x_{oa}\, T_{oa} + (1 - x_{oa})T_{ra}, \qquad x_{oa} = \frac{\dot{m}_{oa}}{\dot{m}_{sa}} \tag{1}$$

The sensible heat delivered by the heating coil follows from the air-side and water-side energy balances, which is why heating output can be regulated through the water flow admitted by the coil valve [5]:

$$\dot{Q}_{hc} = \dot{m}_{sa}\, c_{p,a}(T_{sa} - T_{ma}) = \dot{m}_w\, c_{p,w}\big(T_{w,in} - T_{w,out}\big) \tag{2}$$

At zone level, the simplest control-oriented representation is a single resistance–capacitance (1R1C) node, in which $C_z$ is the effective thermal capacitance of the zone, $R_{env}$ the envelope resistance to outdoors, and the remaining terms are the HVAC, internal and solar heat gains:

$$C_z \frac{dT_z}{dt} = \frac{T_o - T_z}{R_{env}} + \dot{Q}_{hvac} + \dot{Q}_{int} + \dot{Q}_{sol} \tag{3}$$

Higher-order RC networks and data-driven surrogates generalise this structure (Sections 4 and 7). **Figure** 1 shows how the components and feedback loops of the reference plant are arranged.

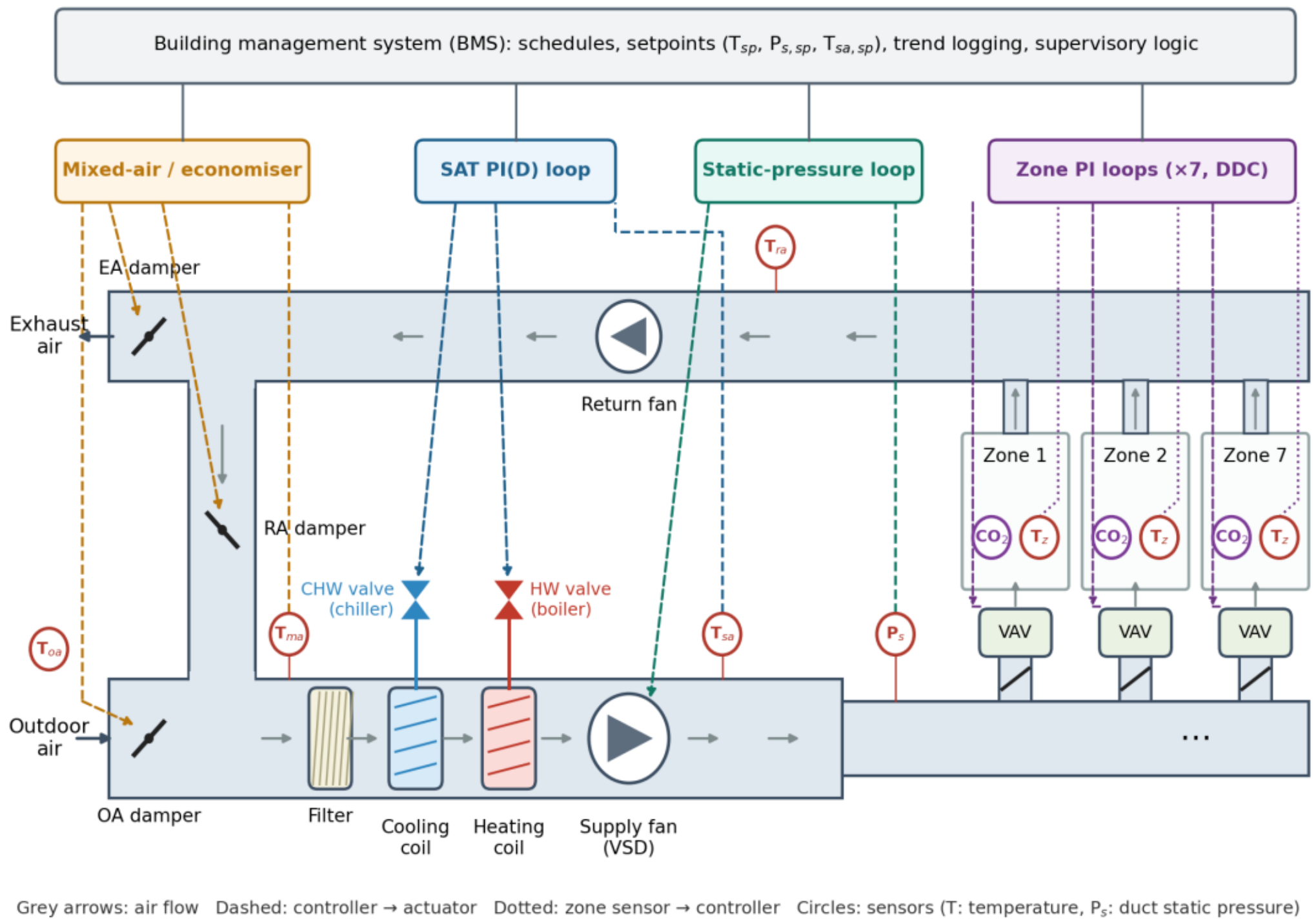

**Figure 1.** Schematic of the reference VAV air-handling plant (configuration after AHU7 in [5]), showing the mixing section, filter, coils, supply and return fans and three of the seven VAV-served zones, together with the four families of feedback loop discussed in Sections 2.1.2–2.1.5: mixed-air/economiser, supply-air temperature (SAT), duct static pressure and zone loops. Sensor set illustrative. Diagram prepared by the authors.

Adjacent VAV boxes may supply a common space, each with its own temperature sensor, when the nature of the space and the airflow patterns cause both sensors to read the same value [9]. Zoning itself is defined as the practice of dividing a building into sections for heating and cooling control such that one controller is sufficient to determine the heating and cooling requirements of the section [8], and where dampers and heaters can be actuated separately in different spaces the thermal zoning can be arranged flexibly, with a single model adapted to the zones that have been defined [10].

### *2.1.2 Hierarchical decomposition of the control system*

The control system of such a plant comprises many control loops and is divided into central-system loops and local- or zone-control loops, all of which should be tied together to share information and system commands through a building management system [8].

The instrumentation-and-control stack is conventionally described as a five-level hierarchy. At Level 0, measurement and control operations are initiated from field devices such as solenoids, sensors, actuators and relays through local plant components as electrical signals; physical signals including temperature, pressure, level and flow are acquired at Level 1 through programmable logic controllers (PLCs) and remote terminal units (RTUs), which interface at Level 2 with local and wide-area network software that enables supervision, monitoring and control by programming devices; Level 3 coordinates the storage of important data and trends, and at Level 4 the top-hierarchy central computer combines data and performs production planning and collective scheduling [11]. The functional elements of the acquisition section — sensors, transducers, transmitters and actuators — collect process variables and transmit them as collective data from which control commands are generated, while the field controllers apply preprogrammed algorithms to produce demand signals, with RTUs and PLCs acquiring scattered data from distributed sensors and transmitting them to the master station [11].

Unitary controllers regulate a single piece of equipment, larger building controllers handle facility-wide programming or monitoring of general input/output points, and PLCs are used in industrial installations, the classes differing in input/output type and capacity, form factor and physical robustness, and processor programming capability [12]. Direct digital control (DDC) unitary controllers are the basic component of most building automation systems: they carry built-in input/output capability and network connections, some have small interface displays, they are made for operating a specific piece of HVAC equipment — a fan-coil unit with a reheat coil, or a cooling-only air handler with an economizer — they accept 24 V alternating current power, and through their network capability they receive plant-wide data such as outdoor air temperature, connect to the supervisory system and respond to data-polling requests from a higher-level controller [12]. A DDC system provides closed-loop control by integrating computers, microprocessors and software with sensors and actuators [13].

### *2.1.3 Instrumentation and actuation*

The measured quantities that feed the feedback loops of the reference architecture are those of a VAV air-handling installation: the supply-air temperature of each thermal zone, the supply-air static pressure, the supply-airflow rate, the supply-water flow rate in summer and in winter, the return-air temperature, the supply and return cooling-water temperatures, the supply and return heating-water temperatures, the ambient temperature, and the temperature after the rotary heat exchanger, all of which serve as inputs to the feedback control loop [6]. The controlled variables named in control-sequence specifications are likewise temperature, pressure, carbon dioxide level, humidity, carbon monoxide level, filter pressure drop, flow and velocity [14].

The ventilation system is controlled based on $CO_2$ levels and exchanges the amount of air required in the rooms, although the exact supply-air temperature cannot be controlled for separate rooms, so that during the cold season the heating system must compensate for the possible temperature drop [10]. The radiators in separate rooms are individually adjustable through valves actuated by the BMS or manually via thermostats, and floor-heating branches can be controlled separately through installed valves [10].

A pneumatically operated steam control valve with actuator and positioner piloted by a controller provides the most precise control at the highest cost; a pilot-operated valve offers a compromise between a direct acting valve and a pneumatic actuated valve and is reasonably priced; and pilot-operated valves are generally the choice for pressure control while pneumatic actuated valves are the choice for temperature control [15]. The responsiveness of any such sensor is characterised by its time constant, defined as the time required for a dynamic component to reach 63.2 percent of its total response to an instantaneous, or step, change to its input [8].

For a first-order sensor or process subjected to a step input whose final response amplitude is $\Delta y_\infty$, the response and its value at one time constant are

$$y(t) = y_0 + \Delta y_\infty\left(1 - e^{-t/\tau}\right), \qquad \frac{y(\tau) - y_0}{\Delta y_\infty} = 1 - e^{-1} \approx 0.632 \tag{4}$$

so that the 63.2 % point of the definition coincides with $t = \tau$.

*2.1.4 Setpoints, feedback signals and command paths*

A zone-control application determines the sets of setpoints for all controlled actuators that, when applied, satisfy comfort and energy-efficiency targets, treating the actuator setpoints as optimisation variables and the comfort and energy targets as objectives, generating a Pareto set of candidate solutions and selecting a final solution from that set by social-welfare criteria [16]. Because the $CO_2$ and temperature dynamics of the controlled zones are comparatively slow, several minutes are available for optimisation between successive actuations, which is what makes a computationally expensive global search admissible [16].

A representative configuration modulates the cooling-coil and heating-coil valves within the rooftop unit to maintain the supply-air temperature at 55 °F year-round, holds the zone air temperature heating setpoint at 69.8 °F during occupancy and reduces it to 60 °F outside occupied hours, the occupied mode running from 7:00 am to 10:00 pm. Comfort targets of this kind are conventionally derived from fixed seasonal values — the ASHRAE-55 standard recommends 20 °C in winter and 24 °C in summer — or from the adaptive comfort model, in which the neutral temperature in °C is obtained from $T_n = 0.31T_{rm} + 17.8$ [2]; PCM-envelope studies have typically presumed standard thermostat settings of 24 °C for cooling and 20 °C for heating irrespective of the storage material [17]. A zero energy band allows temperatures to float between selected settings and thereby prevents the consumption of heating or cooling energy while the temperature lies within that range [8]. Steady-state control logic maintains fixed setpoints for key operational parameters such as cooling-water temperature and flow rates and make-up air unit flow rates, an approach that is unable to respond promptly to rapidly changing outdoor conditions and load fluctuations [18].

In the descriptive stage, data from chillers, pumps and sensors are aggregated in a cloud platform while day-to-day control remains with the existing BMS and no changes are made to the local PID loops, the twin serving purely as a monitoring and diagnostic layer; in the predictive stage the twin supplies supervisory inputs, forecasting conditions and suggesting setpoint adjustments that are applied through the BMS's standard control loops [19]. Field implementations follow the same route: the controller is implemented on top of a standard building automation and control system, and continuous controller decisions are executed by modulating the duty cycle of valve opening in individual rooms, with the thermal power into individual rooms not measured but calculated by dividing the total power of the entire unit by the size of the inlet pipe [20].

Sensor channels monitoring the working condition of the rooftop unit and the VAV boxes — temperature, humidity, pressure, airflow and energy consumption — are logged at a sampling interval of one minute, whereas optimisation is performed over intervals of minutes between actuations [16]. Because the controller's sampling time differs from the simulation timestep, timely communication between the physical system, the digital twin, the controller and the miscellaneous systems is imperative, and middleware is required to manage communication between the functional mock-up units and to synchronise the physical and digital systems [20]. The sampling period is one of the factors that must be taken into account to ensure a stable operating control system, together with correct tuning of controllers, interactions between control loops, and stable operation over all operating conditions such as the variation of system gain with load [9].

#### *2.1.5 Conventions used in the remainder of the review*

Two structural conventions follow from the anatomy above and are retained in the later sections. First, because every input and output signal terminates at a controller, whether a unitary DDC device, a building controller or a PLC [12], any new control algorithm must be specified in terms of the control point, the controlled device and the determination of the setpoint for each loop, together with the control logic — P or PI — any resets, and the direct or reverse acting required for acceptable performance [14]. Second, control loops may interact: in a network of heat emitters fed from a common secondary circuit, the movement of one two-port valve in response to a load change alters the pressure seen by the other circuits, changing their flows and provoking further valve movement, so that the circuits may fail to reach a steady equilibrium and continue to hunt [9]. Control sequences are written by breaking the plant into major subsystems — air handlers, chiller plant and boiler plant, CAV, VAV, reheat coils, exhaust fans, isolation rooms, hoods, air flow monitors — and subdividing each into control blocks for start/stop and for each controlled device or variable, as in the variable-volume example divided into fan start/stop, air velocity control by inlet guide vanes, and temperature and mixed-air control, which interconnect but each operate as separate subsystems [14].

## 3. Thermostat and PID Control: From On/Off Switching to Continuous Regulation

### *3.1 Thermostat and PID Control: From On/Off Switching to Continuous Regulation*

Every higher-level architecture reviewed in this article is defined by what it adds to, or replaces in, the single-loop regulatory layer. This section therefore examines that layer in its own right: the switching thermostat and its deadband, the scheduled setpoint, the proportional–integral–derivative (PID) loop that succeeds them, and the operating conditions under which each of these baseline strategies ceases to perform acceptably.

#### *3.1.1 Two-position control: the differential gap and its costs*

The simplest regulator in the building control repertoire is the two-position, or on/off, controller, in which the controlled device gives either maximum or minimum output, typically on and off [9]. Because a finite temperature differential is necessary between switching the device on and switching it off, the controller operates not at a point but across a band: the temperature interval between the upper and lower limits is termed the differential gap or differential band, known in American usage as the deadband, and within that gap the output may be either on or off depending on the last switching operation [9]. Current convention places the set point at the upper point of the differential gap, whereas earlier conventions take it as the centre point [9]. The immediate consequence is a swing of temperature about the set point — the peak-to-peak variation being termed the swing or operating differential, as distinct from the controller's own mechanical or manual differential — with a mean temperature that normally lies below the set point, although some systems operating under light loads may give a mean above it [9].

For a heating application with the set point at the upper limit of the differential gap, as in the current convention [9], the two-position law can be written as follows, where $\delta$ is the differential and $u(t^-)$ is the output immediately before $t$; the dependence on the previous state is what makes the controller hysteretic:

$$u(t) = \begin{cases} 1, & T_z(t) \leq T_{sp} - \delta \\ 0, & T_z(t) \geq T_{sp} \\ u(t^-), & T_{sp} - \delta < T_z(t) < T_{sp} \end{cases} \tag{5}$$

The deadband is therefore not a free design parameter. Reducing the differential reduces the swing, but only at the cost of increased frequency of switching, with attendant wear on the system [9]. The air-terminal design literature supplies the magnitude of the comfort penalty, stating that two-position systems usually cause a minimum of 2°F (1.1°C) room temperature swings as the thermostat toggles the air valve open and closed [13]. The steam-plant literature records the same trade-off from the actuator side: a boiler or burner that cycles excessively is diagnosed as oversized, as having pressure control settings too close together, or as having too small a steam chest, the consequence being premature burner failure and loss of efficiency [15]. Plant lag aggravates the interaction, since room temperature continues to increase for a time after the heating system has been switched off, for example because of hot water present in the radiators [9].

Hysteresis is not confined to temperature loops. In a documented rule-based ventilation baseline, the variable-air-volume (VAV) damper was stepped through three discrete positions — 45% open at $CO_2$ levels of 750 PPM or less, 70% between 750 and 900 PPM, and 100% above 900 PPM — with a 100 PPM hysteresis applied when $CO_2$ levels are dropping in order to avoid rapid switches around the thresholds [16]. In that installation the maximum $CO_2$ level reached 1046 PPM, above the 1000 PPM threshold of the Danish Working Environment Authority, although the excursion was confined to a very short period to which the ventilation actuation reacted immediately [16]. Banded switching of this kind thus suppresses actuator chatter without eliminating the excursion itself.

### *3.1.2 Scheduled setpoints and the fixed-mode baseline*

The step from a single thermostat to a building-level strategy is normally taken through time scheduling. In the Fixed Mode of a documented hotel installation, chillers, pumps and cooling towers operate on static schedules and predefined temperature setpoints without consideration for real-time feedback from environmental sensors or internal system loads, chillers being activated or deactivated strictly according to time-of-day or fixed temperature thresholds [19]. The same arrangement appears in an experimental facility operated in occupied mode from 7:00 am to 10:00 pm, with a zone heating setpoint of 69.8°F during occupied hours reduced to 60°F during unoccupied hours and a supply air temperature held at 55°F year-round [21]. The mode is reported to produce operational inefficiencies, particularly under part-load conditions or during mild weather, where equipment may run unnecessarily or spaces may be overcooled to maintain conservative safety margins [19].

Simulation and field evidence converge on the existence of a penalty, though not on its measure. Talami et al. compared fixed setpoint control with prediction-based dynamic setpoints and setback strategies under various occupancy patterns and found approximately 15% additional energy savings for dynamic control, particularly in buildings with variable occupancy, achieved by reducing unnecessary HVAC operation during periods of low demand [18]. In occupied buildings, Adhau (2019) reported statistically significant savings of up to 1.5 MWh per day from learning-based model predictive control relative to traditional schedule-based static control while maintaining thermal comfort, and Blum et al. (2022) reported approximately 10% energy savings over a two-month deployment in an office HVAC system relative to a baseline static configuration [18]. The reported figures are expressed in different currencies — percentage savings over a period versus daily absolute energy — and rest on different validation regimes, one simulation-based and two drawn from live deployment [18].

The scheduled setpoint is also a design variable in its own right rather than an attribute fixed by the building. Prior work has frequently presumed standard thermostat settings, such as 24°C for cooling and 20°C for heating, regardless of the characteristics of the phase change material (PCM) in the envelope, an assumption argued to hide the full potential of latent thermal storage, and Saffari et al. have emphasized that the ideal interaction between PCM envelopes and HVAC controls is a largely unexamined research domain, particularly regarding energy flexibility and demand response [17]. Synchronizing HVAC operation schedules with the thermal characteristics of PCMs, for instance by facilitating pre-cooling or thermal drift, is reported to enhance the latent storage advantages of PCMs without sacrificing thermal comfort, and NSGA-II has been used to optimize insulation levels and thermostat setpoints simultaneously, yielding improvements in comfort and energy efficiency across Iranian climate zones [17]. The economics of the device itself have been assessed for adaptive thermostats, whose costs are reported as recoverable in less than two years, with the accompanying recommendation that recommended insulation thickness values in building codes and guidelines be revised when adaptive thermostats are employed for indoor temperature control [2].

### *3.1.3 From proportional action to PI and PID*

Continuous regulation replaces the switching decision with a position. In proportional control the final control element moves to a position proportional to the deviation of the controlled variable from the setpoint, the control-point range through which the controlled variable must pass to move that element through its full operating range being the throttling range or proportional band, expressed in values of the controlled variable [8]. Modulating systems, which allow control settings over a wide range between fully open and fully closed, typically provide more stable conditions than two-position systems [13]. Proportional action alone nevertheless cannot remove the residual error: with strictly proportional action the room variance from set point is maintained throughout the occupied period, so that something must be added to the proportional device to keep the occupied space at or near set point [13].

In the notation adopted here, with error $e = r - y$, the proportional law and its band are given below, where $u_0$ is the output at zero error (the manual reset or bias) and $PB$ is the proportional band expressed as a percentage of span when $u$ and $y$ are normalised [85]:

$$u(t) = u_0 + K_p\, e(t), \qquad e(t) = r(t) - y(t), \qquad PB = \frac{100}{K_p}\ \% \tag{6}$$

For a self-regulating plant of steady-state gain $K$, a step $\Delta r$ in the reference — or an equivalent step of load referred to the output — leaves the residual error [85]

$$e_{ss} = \frac{\Delta r}{1 + K\, K_p} \tag{7}$$

which falls as $K_p$ increases but vanishes only in the limit $K_p \to \infty$; because raising the gain erodes the stability margin, the offset cannot be tuned away, which is the quantitative content of the observation that proportional action alone cannot hold the space at set point [13].

Proportional-integral (PI) control supplies that addition, combining the proportional response with an integral, or reset, response that tends to correct the offset resulting from proportional control [8]. In the terminal-unit formulation the integral factor amplifies the proportional offset and incorporates a timing function that increases or decreases the integral contribution if room temperature is not moving toward set point at an acceptable rate, converting the thermostat's temperature signal into a demand signal and decreasing the time for the terminal unit to recapture set point, with the integral contribution large when the error is large and small when the error is small [13]. PI control is reported to be available with many pneumatic and analog electronic controls and virtually standard on digital control systems [22], and most HVAC control loops are described as at least PI controlled [13]. On where field practice settles, the two literatures agree.

They diverge on the third term. The design guidance presents proportional plus integral plus derivative (PID) control as the tightest of the three options, controlling the room to less than 0.5°F (0.28°C) or better depending on the derivative setting, against a variation of less than 2°F (1.1°C) for PI depending on the integral setting and a minimum of 2°F (1.1°C) swing for two-position control [13]; the definitional literature likewise credits the derivative component with compensating for system dynamics and allowing a faster control response [8]. Field-oriented guidance takes a materially different view of its worth: adding the derivative term reduces overshooting and applies "brakes" to overreacting integral terms, and derivative control has a very fast response that suits fast-acting industrial processes and rocketry, but because most HVAC system responses are relatively slow the value of derivative control in most HVAC applications is minimal, the term may complicate the tuning process and cause unstable responses, and derivative logic is therefore normally not used in most field HVAC applications [22]. The same source observes that the generic control loop is often referred to as a PID loop, particularly in digital control applications, even though the derivative function is typically unused [22]. The label PID consequently does not reliably denote the presence of derivative action in building applications, and the reported sub-0.5°F performance should be read as a property of a genuinely derivative-weighted loop rather than of typical practice. The performance intervals themselves overlap, moreover, since the PI figure of less than 2°F (1.1°C) sits at the floor of the two-position range, so the advantage of PI over on/off is contingent on the integral setting rather than structural [13].

The standard (ideal, non-interacting) form of the PID law and its parallel-gain equivalent are [85]

$$u(t) = K_p \left[ e(t) + \frac{1}{T_i} \int_0^t e\,(\xi)\, d\xi + T_d \frac{de(t)}{dt} \right], \qquad K_i = \frac{K_p}{T_i}, \quad K_d = K_p\, T_d \tag{8}$$

Digital controllers execute the law at a sampling interval $\Delta t$; a widely used implementation is the incremental or velocity form, obtained by differencing the backward-Euler discretisation of Eq. (8) [85]:

$$\Delta u_k = K_p \left[ (e_k - e_{k-1}) + \frac{\Delta t}{T_i} e_k + \frac{T_d}{\Delta t} (e_k - 2e_{k-1} + e_{k-2}) \right] \tag{9}$$

Because only the increment $\Delta u_k$ is computed, the actuator position itself carries the integral state, which simplifies bumpless transfer between manual and automatic operation [85].

**Figure** 2 contrasts the four strategies on a common illustrative zone model of the first-order-plus-dead-time type (Eq. (11), Section 3.1.5; $K = 20$ K, $\tau = 90$ min, $\theta = 3$ min) subjected to a 3 K drop in outdoor temperature followed by a step in internal gains. The simulation reproduces the qualitative ordering reported in the design literature [13]: the on/off loop cycles with a peak-to-peak swing of about 1.7 K and a mean below set point, the proportional loop settles with exactly the offset predicted by Eq. (7) (0.33 K for $K\,K_p = 8$), and the PI and PID loops remove the offset, with peak deviations after the outdoor-temperature drop of about 0.19 K and 0.13 K, respectively, for the gains shown.

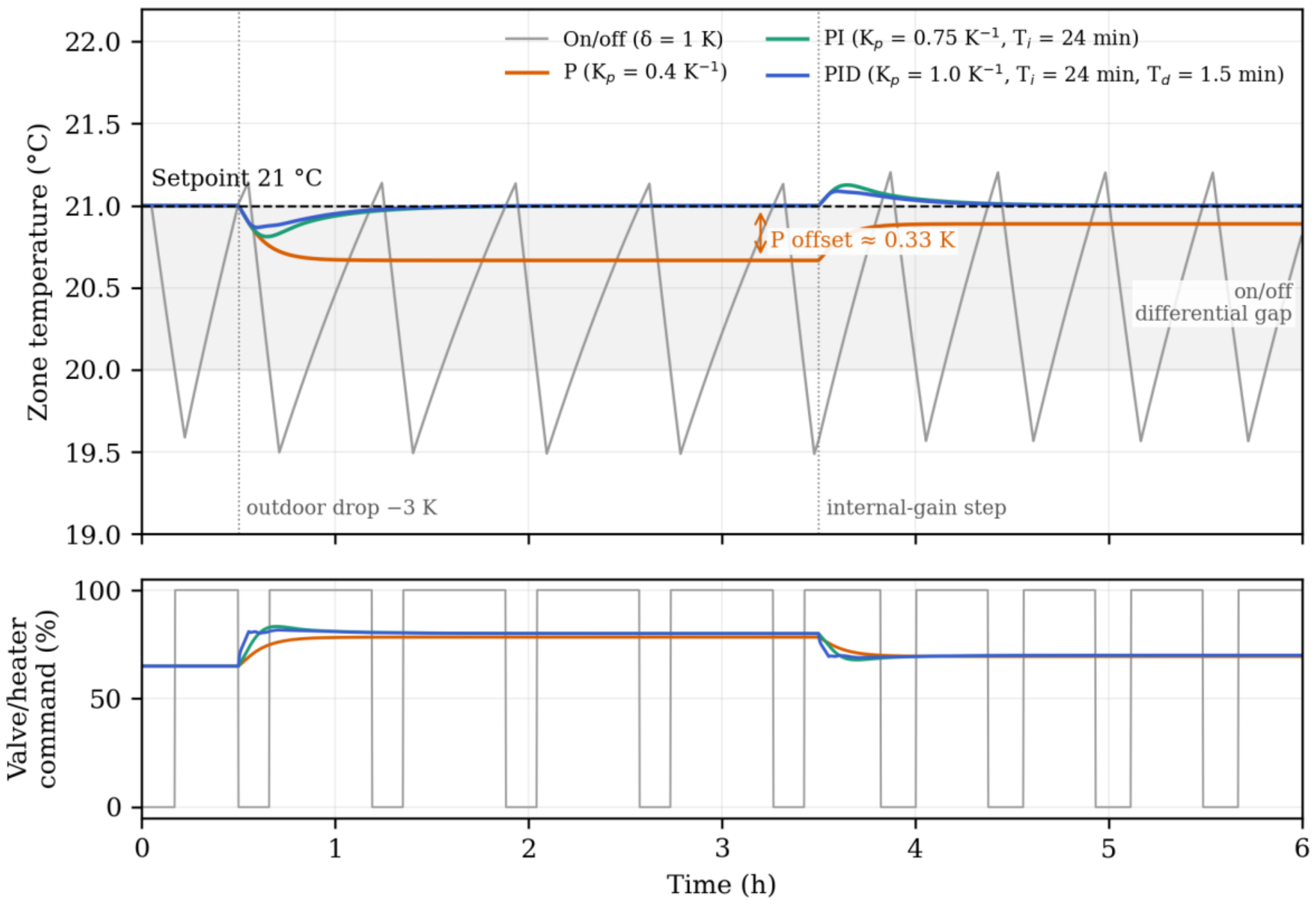

**Figure 2.** Simulated zone-temperature responses (top) and actuator commands (bottom) under two-position (on/off, 1 K differential gap with the set point at its upper limit), P, PI and PID control of an illustrative first-order-plus-dead-time heated zone ($K = 20$ K, $\tau = 90$ min, $\theta = 3$ min, sampling 10 s). Disturbances: outdoor temperature −3 K at 0.5 h; internal-gain step equivalent to +2 K at 3.5 h. The P offset equals $\Delta/(1 + KK_p)$ as in Eq. (7). Simulation by the authors.

### *3.1.4 Windup, derivative action and mode switching*

The most frequently documented pathology of integral action in buildings is windup. It arises when the control loop is operating but the controlled device is disconnected or otherwise unable to control the controlled variable, such as when a system is turned off at night: the variable cannot be maintained at set point, the integral term becomes larger and larger, and when the system is turned on the manipulated variable sits fully in one direction so that the system will usually overshoot the set point, the integral term taking time to fall because of the long period during which the system was far from set point, with the effect of rendering the system temporarily unstable [22]. The mitigation hierarchy is explicit: disabling the controller when the system is turned off is the preferred solution, with adding derivative control or applying one of the anti-windup devices and algorithms commonly used with analog electronic controllers as alternatives, a common anti-windup algorithm being to use proportional logic only until the system has been on for a period of time [22]. Schedule-driven operation of the kind described above — an occupied heating setpoint of 69.8°F reduced to 60°F during unoccupied hours [21] — supplies precisely the periodic off-condition in which windup is generated [22]. There is an internal tension in this guidance, since derivative action is offered as one remedy for windup in the same treatment that concludes derivative logic is normally not used in field HVAC applications [22].

Anti-windup schemes act on the integrator state once the actuator saturates. Writing $v$ for the unsaturated controller output and $u$ for the applied command, the back-calculation scheme feeds the saturation error back into the integrator through a tracking time constant $T_t$, so that the integral $I$ stops growing as soon as $u \neq v$ [85]:

$$u = \mathrm{sat}(v) = \min(\max(v, u_{min}), u_{max}), \qquad \frac{dI}{dt} = \frac{K_p}{T_i} e(t) + \frac{1}{T_t}(u - v) \tag{10}$$

Conditional integration, which freezes $I$ while the output is saturated, and disabling or resetting the controller while the plant is off — the remedy that [22] identifies as preferred — are the common alternatives. **Figure** 3 illustrates the mechanism for a night shutdown during which the loop remains active: without protection, the integral accumulated over eight hours holds the valve fully open after restart and the zone overshoots by more than 6 K without recovering within the simulated period, whereas back-calculation and disabling the controller while the plant is off both return the zone to set point with an overshoot below 0.5 K.

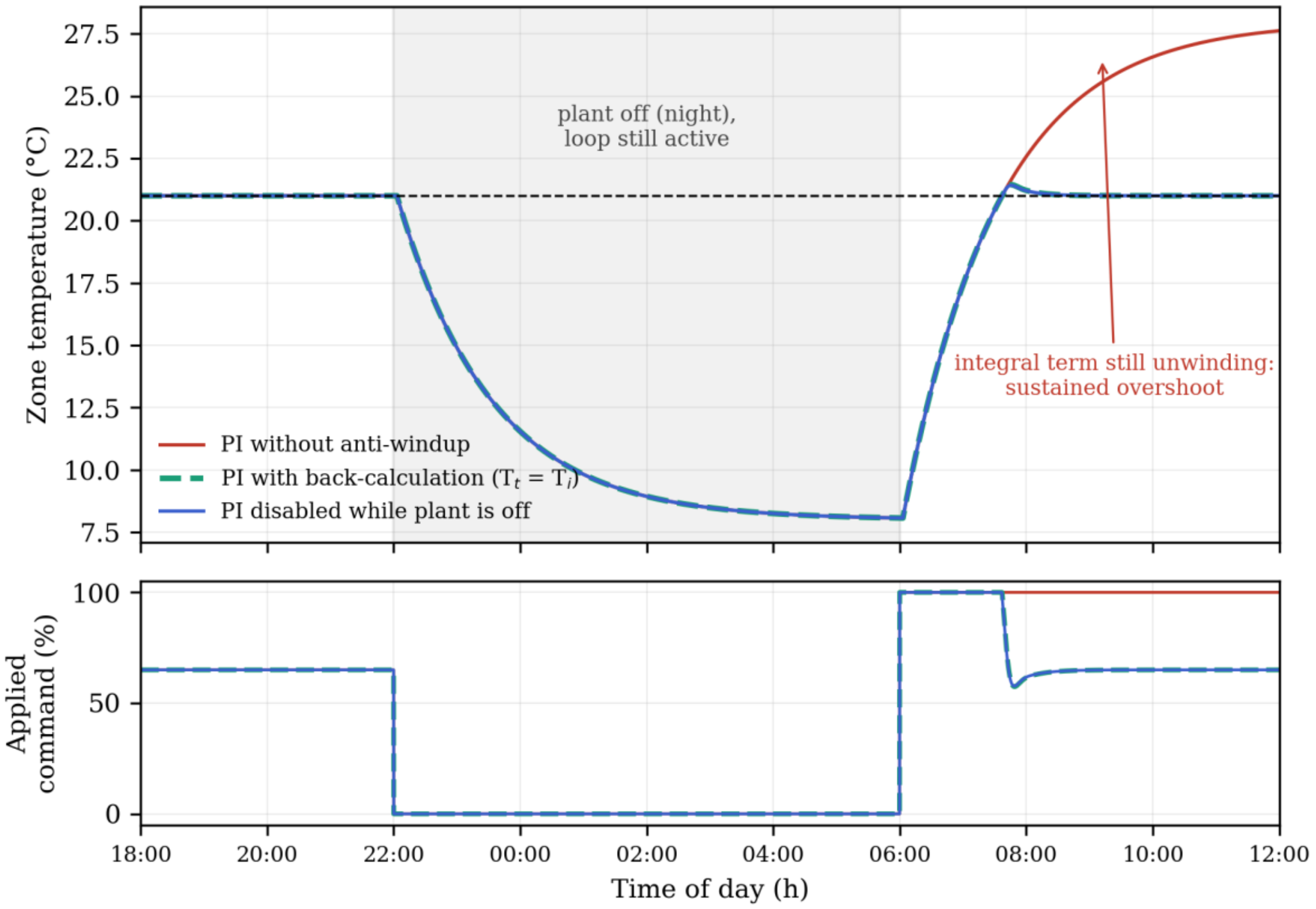

**Figure** 3. Integral windup during a night shutdown (22:00–06:00) with the PI loop left active at a 21 °C set point ($K_p = 0.75$ K$^{-1}$, $T_i = 24$ min; same zone model as **Figure** 2). Without anti-windup the integrator saturates the valve after restart and causes a sustained overshoot; back-calculation (Eq. (10), $T_t = T_i$) and disabling the controller while the plant is off give practically identical, well-damped recoveries (curves overlap). Simulation by the authors.

A second discontinuity is the change of operating mode itself. Discrete switching arising from varying setpoints or operating modes is one of the characteristics that, together with unpredictable stochastics from user interaction, external and internal dynamics from weather conditions and internal loads, and asynchronous or subcomponent interacting operations from parallel component structure, makes the modelling of hybrid nonlinear dynamic systems challenging and renders the derivation of universally applicable white-, black- or grey-box modelling schemes difficult [23]. Migration away from switching is nonetheless reported outside air-side loops: in a steam-generator context, on/off pressure switches are advised to be replaced with PID-type pressure controls, which give additional flexibility for burner response to system pressure [15].

*3.1.5 Tuning rules and their transferability to building loops*

Tuning is where the classical literature and building practice are least aligned. Conventional PID tuning approaches are categorized into two groups: rule-based approaches, such as Ziegler–Nichols and Cohen–Coon, and model-based approaches, such as internal model control (IMC) [3]. The applicability of both families is bounded by the assumption of a fixed plant, since most practical systems in industry are nonlinear and time-varying, which makes the conventional approaches less applicable because they would require frequent re-tuning whenever there are changes in the system — a difficulty that becomes more prominent when implementing PID control for complex systems such as nonlinear multiple-input, multiple-output (MIMO) systems under time-varying uncertainty [3]. That assessment is framed for industrial plant in general rather than for building loops specifically [3]. Building-specific guidance frames tuning instead as the search for the best combination of response speed and stability, and points forward to adaptive controls that learn by experience how to optimise the operation of a controller [9].

Both families are usually stated for the first-order-plus-dead-time (FOPDT) approximation of the process, with steady-state gain $K$, time constant $\tau$ and apparent dead time $\theta$ — the model class that also underlies the pattern-recognition adaptive controller discussed in Section 7.1.4 [9]:

$$G(s) = \frac{Y(s)}{U(s)} = \frac{K\,e^{-\theta s}}{\tau s+1} \tag{11}$$

The closed-loop Ziegler–Nichols rule needs only the ultimate gain $K_u$ at which the proportional loop oscillates with constant amplitude and the corresponding period $P_u$, and sets the PID parameters to [85]

$$K_p = 0.6\,K_u, \qquad T_i = \frac{P_u}{2}, \qquad T_d = \frac{P_u}{8} \tag{12}$$

its proportional-only setting of $0.5\,K_u$ is the formal counterpart of the field practice of raising the gain until oscillation and backing off to about one-half of that value [22]. Tuning quality is commonly scored by integral error criteria; the integral of time-weighted absolute error over an evaluation window $T_w$ is the measure used in the deep-RL tuning study discussed in Section 5.1.2 [3]:

$$\mathrm{ITAE} = \int_0^{T_w} t\,|e(t)|\,dt \tag{13}$$

The design guides quantify what different settings purchase without specifying a procedure for reaching them, since achievable room variation is stated as a function of the integral setting for PI and of the derivative setting for PID [13]. Where an explicit adaptive mechanism is required, two named schemes are reported: model reference adaptive control (MRAC), which compares the system's actual output with that of a desired reference model and adjusts control parameters to minimise the error using a mathematical model of the system to estimate its dynamics, and self-tuning control (STC), which continuously estimates the system's dynamics and adjusts parameters on the basis of observed behaviour, typically through identification methods such as recursive least squares or instrumental variable methods [24]. The applications cited for these techniques are aerospace, robotics, manufacturing, chemical plants and power systems rather than building HVAC [24]. At the level of actual implementation, a PID loop in a test-house temperature control system was implemented in the control software and actuated through a solid-state relay, regulating heater power output to hold the house at a set temperature of 20°C [25].

### *3.1.6 Failure modes that motivate the remainder of the review*

Four characteristics of building plant are consistently identified as the limits of the thermostat–PID layer. The first is nonlinearity and time-variance, which undermines rule-based and model-based tuning alike and is most acute in MIMO systems under time-varying uncertainty [3]. The second is discrete mode switching, whose effect on the tractability of hybrid nonlinear dynamic models is noted above and which no fixed-gain loop anticipates [23]. The third is loop interaction at the multi-zone scale. Cleanroom guidance states the principle as an injunction against controlling to the same parameters in series: controlling the air handler and the zones by flow causes unstable loops that may fight for control, whereas controlling the air handler discharge by duct static pressure and the zones by airflow exploits two measurements with different dynamic characteristics that are less likely to result in unstable control interferences, a consideration that becomes more important with complicated systems serving many zones simultaneously [12]. The general statement of the same difficulty is the nonlinear MIMO system under time-varying uncertainty for which adaptive PID tuning was developed [3]. The fourth is the timescale of the plant itself: because most HVAC responses are relatively slow, the fast-response capability of derivative action contributes little, while the fixed schedule under which the loop operates takes no account of real-time feedback from environmental sensors or internal system loads and is reported to leave equipment running unnecessarily or spaces overcooled under part-load and mild-weather conditions [19, 22].

Read together, these findings describe a regulatory layer that is inexpensive, well codified and adequate for holding a single variable against a fixed target, but one whose target is set outside the loop and whose gains are matched to a plant assumed constant. Even the interaction between the scheduled setpoint and the thermal storage embedded in the envelope remains, in Saffari et al.'s assessment, a largely unexamined domain [17]. The architectures examined later in this article are most usefully read as responses to precisely these four conditions.

**Table 1.** Thermostat-level and PID strategies: governing principle, reported performance, and recorded trade-offs.

| **Strategy** | **Governing principle** | **Reported control performance** | **Advantages and available mitigations** | **Limitations** |
|---|---|---|---|---|

| Strategy | Governing principle | Reported control performance | Advantages and available mitigations | Limitations |
|---|---|---|---|---|
| Two-position (on/off) | Controlled device gives maximum or minimum output; switching separated by a differential gap (deadband), with the set point at the upper gap limit under current convention [9] | Minimum 2°F (1.1°C) room temperature swings as the thermostat toggles the air valve [13] | Wider pressure settings or burner derating reduce cycling [15]; hysteresis avoids rapid switching around thresholds [16] | Swing reducible only at the cost of switching frequency and wear [9]; excessive cycling produces premature burner failure and loss of efficiency [15]; mean temperature normally below set point [9] |
| Proportional (P) | Final control element position proportional to the deviation from setpoint, over a throttling range or proportional band [8] | Modulating control typically gives more stable conditions than two-position control [13] | Stable modulating action in place of switching [13] | Room variance from set point maintained throughout the occupied period; an additional term is required to reach set point [13] |
| PI | Proportional response combined with integral (reset) response that corrects the offset of proportional control [8] | Room variation less than 2°F (1.1°C), depending on the integral setting [13] | Integral (reset) action removes the proportional offset [8]; virtually standard on digital control systems [22]; most HVAC loops are at least PI controlled [13] | Integral windup when the controlled device cannot act, e.g. night shutdown, causing overshoot on restart [22]; advantage over on/off contingent on the integral setting [13] |

# 4. Reinforcement Learning Formulation for Building Control

## *4.1 Reinforcement Learning Formulation for Building Control*

### *4.1.1 The control loop as a Markov decision process*

Reinforcement learning (RL) is an approach that employs algorithms or "agents" to navigate sequential decision making in an environment with limited feedback, and is consequently well suited to defining policies and behaviours for building control [4]. In the formulation adopted here, the agent, the environment, the states, the actions and the reward signals are the principal constituents, and the Markov decision process (MDP) supplies the mathematical representation that supports decision making in an environment where uncertainty exists [11]. The agent acts, receives feedback through rewards or penalties, and updates its control strategy so as to maximise the cumulative reward [24]; this interactive, trial-and-error learning of an optimal policy is what separates RL from supervised learning, which relies on input–output pairs, and from unsupervised learning, which operates without labelled data [26].

Formally, an MDP is the tuple $(\mathcal{S}, \mathcal{A}, P, R, \gamma)$, where $\mathcal{S}$ and $\mathcal{A}$ are the state and action spaces, $P(s' \mid s, a)$ is the probability of moving to state $s'$ when action $a$ is taken in state $s$, $R$ is the reward function and $\gamma$ the discount factor [86]. At each decision step $t$ the agent observes $s_t$, selects $a_t$ from its policy $\pi(a \mid s)$ and receives the reward $r_{t+1}$ as the plant moves to $s_{t+1}$; its objective is the expected value of the discounted return [86]

$$G_t = \sum_{k=0}^{\infty} \gamma^k \, r_{t+k+1}, \qquad 0 \leq \gamma < 1 \tag{14}$$

The quality of a policy is expressed by its state-value and action-value functions [86],

$$V^{\pi}(s) = \mathbb{E}_{\pi}[G_t \mid s_t = s], \qquad Q^{\pi}(s,a) = \mathbb{E}_{\pi}[G_t \mid s_t = s, \; a_t = a] \tag{15}$$

and the optimal action-value function satisfies the Bellman optimality equation, from which an optimal policy follows by acting greedily [86]:

$$Q^{*}(s,a) = \mathbb{E}[r_{t+1} + \gamma \max_{a'} Q^{*}(s_{t+1}, a') \mid s_t = s, \; a_t = a], \qquad \pi^{*}(s) = \mathrm{argmax}_a Q^{*}(s,a) (16)$$

**Figure** 4 places these quantities on the building control loop.

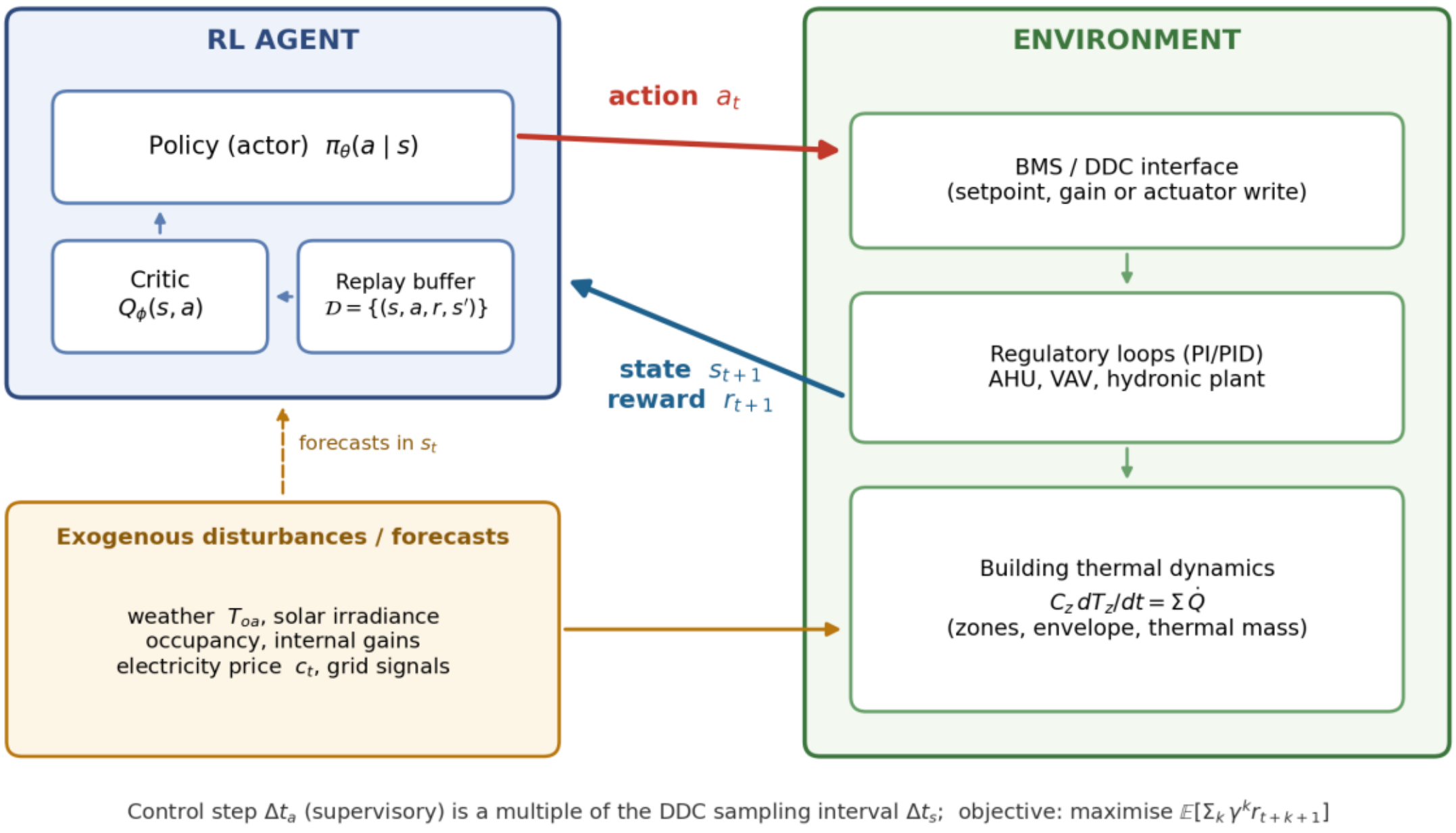

**Figure 4.** The reinforcement-learning control loop for building HVAC. The agent (an actor–critic configuration with replay memory is shown) writes its action through the BMS/DDC interface; the regulatory loops and the building thermal dynamics form the environment, which returns the next state and the reward. Weather, occupancy and price act on the environment and, as forecasts, may enter the state vector. Diagram prepared by the authors on the basis of the formulation in [86].

Two departures from the textbook setting matter for buildings. First, the Markov property holds only approximately: the thermal state of the envelope and of the hydronic network is observed only partially through zone and supply-temperature sensors, while weather, occupancy and tariffs are exogenous processes, so the state vector is usually augmented with lagged measurements, calendar features and forecasts to bring the observed process close to Markovian. Second, the agent's decision interval $\Delta t_a$ is a supervisory quantity, distinct from — and usually much longer than — the sampling interval of the DDC loops beneath it (Section 2.1.4). Reviews of the field organise RL building-control studies around the same constituents that appear in the formal definition: Wang and Hong decompose each study into its algorithm, state, action, reward and environment and catalogue the variants used for each [87], while Vázquez-Canteli and Nagy, reviewing RL for demand response, find that most articles address single-agent systems in stationary environments [88].

*4.1.2 State, action and reward design for HVAC*

The state vector encodes what the agent needs in order to predict the consequences of its actions. In the multi-zone VAV formulation of Wei et al., it comprises the current time, the zone temperatures and the environmental disturbances — outdoor temperature and solar irradiance — and the action is the conditioned-air flow rate supplied to each zone [89]. The action variable, rather than the algorithm, determines where the agent attaches to the plant: Brandi et al. let the agent write the supply-water temperature setpoint for the terminal units of an office heating system [90], whereas Biemann et al. evaluated agents that manipulate the temperature setpoints and fan air mass flow rates of a two-zone data centre simulated in EnergyPlus [91]. Discrete action sets suit value-based algorithms, while setpoints and flows are naturally continuous and favour actor–critic methods (Section 4.1.3), a distinction already noted for continuous energy-management actions in [29].

The reward converts the competing objectives of energy, cost and comfort into a single scalar. A representative form combines an energy-cost term with a comfort-violation penalty summed over the set of zones $\mathcal{Z}$, where $P_t$ is the HVAC electrical power, $c_t$ the time-varying tariff, $\Delta t_a$ the decision interval and $w_E$, $w_C$ non-negative weights:

$$r_t = -\left[w_E\, c_t\, P_t\, \Delta t_a + w_C \sum_{z\in\mathcal{Z}} \left(\left[T_{z,t} - T_{up}\right]^+ + \left[T_{low} - T_{z,t}\right]^+\right)\right], \qquad [x]^+ = \max(0, x) \tag{17}$$

Wei et al. use a reward of this structure, combining a zone-temperature-violation term with an estimate of each zone's contribution to the energy cost under a time-of-use tariff [89]. The comfort band may be fixed, as with the seasonal ASHRAE 55 setpoints, or adaptive, in which case it is centred on the neutral temperature of the adaptive comfort model introduced in Section 2.1.4 [2], with $\Delta_c$ the half-width of the acceptability band adopted:

$$T_{low} = T_n - \Delta_c, \qquad T_{up} = T_n + \Delta_c, \qquad T_n = 0.31\, T_{rm} + 17.8 \tag{18}$$

The ratio $w_E/w_C$ selects an operating point on the energy–comfort trade-off. Because the agent maximises the reward exactly as written, the weighting inherits the transferability problem already noted for aggregated MPC cost functions [16], and Biemann et al. accordingly verified their findings under different reward functions as well as different weather conditions [91].

An alternative to weighting is to remove a requirement from the reward altogether and enforce it as a hard constraint outside the agent. In a direct PPO-based controller for an air-handling unit, trained in a self-contained Python environment that couples a second-order (2R2C) thermal model with a dynamic $CO_2$ mass balance, the reward penalises only the squared deviation of the zone temperature from a 22 °C setpoint, while a hierarchical flow logic overrides any agent action that would drive indoor $CO_2$ above 1000 ppm and an enthalpy-based economiser supplies free cooling from outdoor air [95]. Indoor air quality is thereby treated as a primary constraint rather than as one penalty among others, which removes one weight from the trade-off and guarantees the ventilation requirement irrespective of the learned policy.

*4.1.3 Solution methods: value-based and actor–critic algorithms*

Value-based methods estimate $Q^*$ directly. Tabular Q-learning replaces the expectation in Eq. (16) with sampled transitions and updates its estimate with a step size $\alpha$, converging to $Q^*$ for finite MDPs under standard conditions on exploration and step size [86]:

$$Q(s_t, a_t) \leftarrow Q(s_t, a_t) + \alpha[r_{t+1} + \gamma \max_a Q(s_{t+1}, a) - Q(s_t, a_t)] \tag{19}$$

Building states are continuous and high-dimensional, which is what the deep Q-network (DQN) of Mnih et al. addresses: a neural network $Q(s, a; \theta)$ is trained by minimising the temporal-difference loss over transitions sampled from a replay memory $\mathcal{D}$, with the bootstrap target computed by a periodically updated copy $\theta^-$ of the network [92]:

$$\mathcal{L}(\theta) = \mathbb{E}_{(s,a,r,s') \sim \mathcal{D}}\left[\left(r + \gamma \max_{a'} Q(s', a'; \theta^-) - Q(s, a; \theta)\right)^2\right] \tag{20}$$

Applied to a multi-zone VAV building co-simulated with EnergyPlus, a DQN controller achieved energy-cost reductions of 20–70 % relative to a rule-based baseline while keeping zone temperatures within the desired range [89].

Because the maximisation in Eqs. (16)–(20) runs over a discrete action set, continuous setpoints and flows are handled more naturally by policy-gradient and actor–critic methods, which move the parameters $\theta$ of a stochastic policy along the direction given by the policy-gradient theorem, with $Q$ estimated by a learned critic [86]:

$$\nabla_\theta J(\theta) \propto \mathbb{E}_{\pi_\theta}[\nabla_\theta \ln \pi_\theta(a \mid s)\, Q^{\pi_\theta}(s, a)] \tag{21}$$

Soft actor–critic (SAC) augments the return with an entropy bonus weighted by a temperature $\lambda_H$, where $\rho_\pi$ is the state–action distribution induced by $\pi$ and $\mathcal{H}$ the policy entropy; the bonus sustains exploration and, combined with off-policy learning from a replay memory, improves sample efficiency [93]:

$$J(\pi) = \sum_t \mathbb{E}_{(s_t, a_t) \sim \rho_\pi}\left[r(s_t, a_t) + \lambda_H\, \mathcal{H}\left(\pi(\cdot \mid s_t)\right)\right] \tag{22}$$

Biemann et al. compared four actor–critic algorithms — SAC, TD3, PPO and TRPO — on continuous HVAC control in a simulated data centre and found that SAC reduced energy consumption by about 15 % while maintaining comfort, reaching stable performance with roughly ten times less data than the on-policy methods [91].

*4.1.4 Training environments, deployment evidence and open barriers*

Because exploration on an occupied building is costly, agents are usually trained against a simulator — EnergyPlus co-simulation in [89] and [91], an integrated simulation environment of an office building in [90] — before deployment. Brandi et al. tested a trained agent under four scenarios that varied weather, occupancy patterns and indoor-temperature setpoint requirements, and obtained heating-energy savings of 5–12 % together with improved indoor temperature control, under both static and dynamic deployment [90]. The step to a physical building is documented by Silvestri et al., who developed a SAC agent for the thermally activated building system of a real building using a simplified resistance–capacitance model, of the type in Eq. (3), calibrated on measured data [94]. In simulation the agent matched an ideal MPC in temperature control while consuming 29 % more energy; in the building, over a two-month cooling season, it improved indoor temperature control by 68 % relative to the best-performing rule-based controller without increasing energy consumption [94].

These results frame the barriers taken up in the remaining sections. Wang and Hong found that only 11 % of the RL building-control studies they reviewed had been applied in real buildings, and identified three obstacles to adoption: training that is time-consuming and data-demanding, control security and robustness that need to be strengthened, and limited generalisation that calls for approaches such as transfer learning [87]; Vázquez-Canteli and Nagy likewise report that RL controllers had been tested on physical systems in only a small fraction of the studies they reviewed [88]. The first obstacle motivates the simulators and digital twins of Section 7, the second the constraint-handling and safety layers of Sections 5 and 8.1.2, and the third the transfer-learning agenda of Section 8.1.3. The formulation also fixes the question addressed next: the MDP leaves open which variable of the plant the action $a_t$ denotes, and that choice — gain, actuator demand, reference or setpoint — is what distinguishes the architectures of Section 5.

## 5. Where RL Meets PID: A Taxonomy of Control Architectures

### *5.1 Where RL Meets PID: A Taxonomy of Control Architectures*

#### *5.1.1 The invariant loop and the locus of learning*

Any taxonomy of learning-based HVAC control must begin with the object that does not change. In the classical single-loop formulation, the error signal $x_e$ is processed by a controller that produces an output $Y$ driving a controlled device — an actuator and a fan coil, for example — and hence a change in the controlled variable $x$ [9]. Even where the components of the controlled device have been matched to linearise heat output, the gain of that system can change with operating conditions [9]. The hardware that closes the loop in commercial plant is the direct digital control (DDC) unitary controller, which carries built-in input/output capability and network connections, is often preprogrammed for a specific application such as a fan-coil unit with a reheat coil or a cooling-only air handler with an economiser, and typically accepts 24 V alternating current [12]; field controllers of this class generate control commands from sensor or actuator inputs using preprogrammed algorithms [11]. Any learning architecture, whatever its internal machinery, must ultimately write to some terminal of this chain.

The four classes distinguished below differ, first and most simply, in the variable the learning agent emits: a gain vector for the existing loop, an actuator demand replacing the regulatory law, a reference trajectory computed upstream of the loop, or a setpoint and schedule that defines that reference. This is an architectural rather than an algorithmic distinction, and it cuts across the choice of learning method. It is also a distinction that the literature has tended not to make explicit, in part because the two communities formulate the same problem in incompatible notation: Arroyo et al. observe that "the control and machine learning communities keep evolving independently with a radically different notation natively adopted to formulate the same problem" when comparing model predictive control (MPC) with data-driven control [4]. The performance band that the regulatory layer already delivers sets the benchmark against which each class must be judged: two-position control typically produces room-temperature swings of at least 2 °F (1.1 °C), proportional-plus-integral (PI) control holds the room to a variation below 2 °F (1.1 °C) depending on the integral setting, and proportional-plus-integral-plus-derivative (PID) control holds it to less than 0.5 °F (0.28 °C) or better depending on the derivative setting, with most HVAC control loops at least PI controlled [13].

**Figure** 5 locates the four classes on the layered control stack.

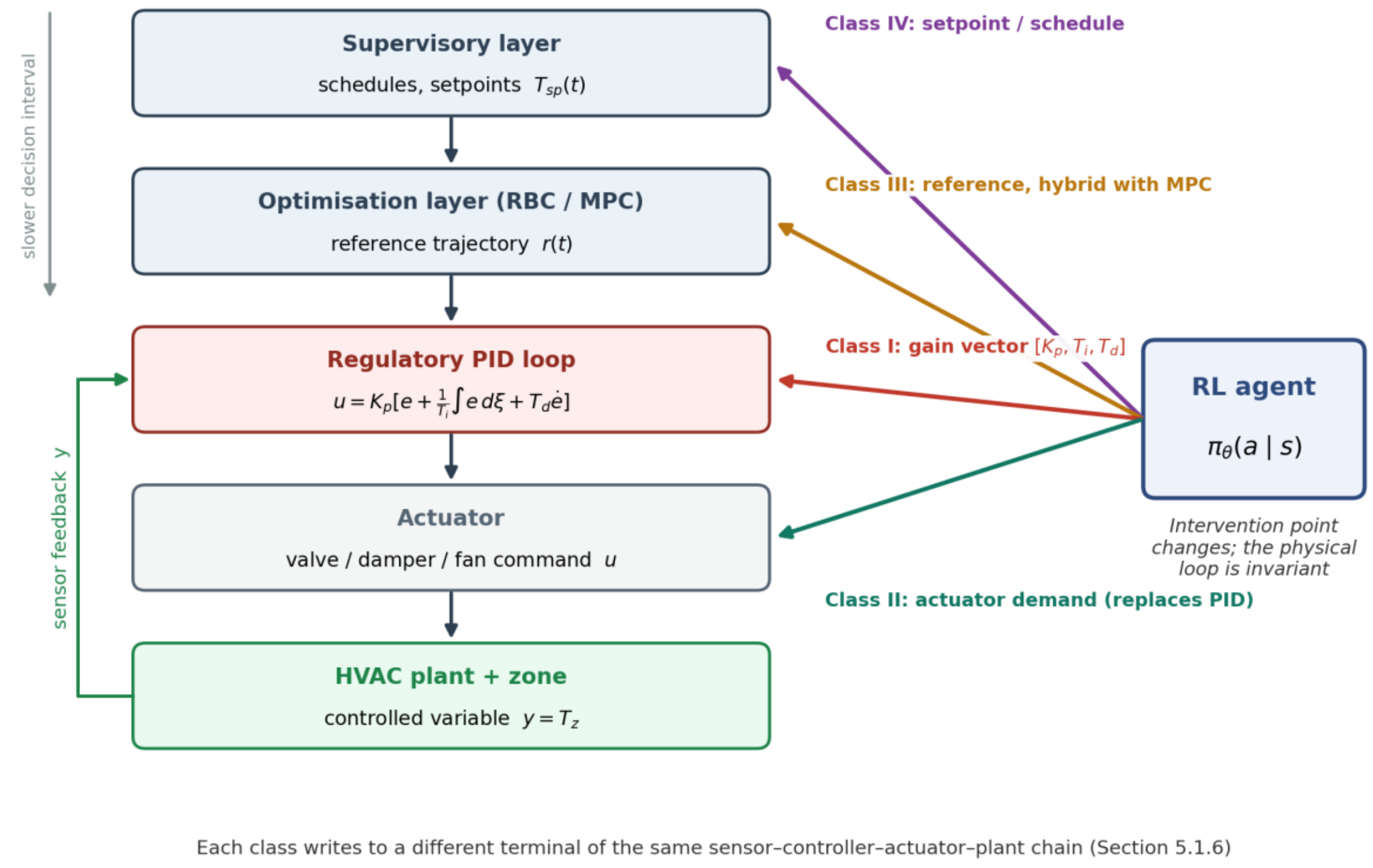

**Figure 5.** Taxonomy of RL–PID control architectures as points of attachment on one layered control stack. Class I writes the gains of the existing PID loop, Class II replaces the regulatory law and writes the actuator demand, Class III computes the reference jointly with an optimisation (MPC) layer, and Class IV writes the setpoint and schedule. Diagram prepared by the authors.

### *5.1.2 Class I — RL as gain tuner for an existing PID loop*

In the first architecture the regulatory law is left intact and the agent acts only on its parameters. The agent observes the loop's response and returns proportional, integral and derivative gains, which the PID block then applies; the closed-loop error continues to be formed by the conventional controller, and the agent never issues an actuator command. The clearest formulation of this architecture is a deep reinforcement learning (RL) tuning strategy for a class of nonlinear multiple-input–multiple-output (MIMO) systems with time-varying uncertainty, in which the agent's own state vector, action vector and reward function are designed so that the resulting policy performs real-time adaptive MIMO PID tuning while the time-varying uncertainty is incorporated into agent training. In a case study on a nonlinear MIMO continuous stirred-tank reactor with two time-varying catalyst scenarios, deactivation at $\alpha_{c,d}$ = 0.0034 and regeneration at $\alpha_{c,r}$ = −0.00334, conventional PID controllers failed to address the time-varying uncertainty and produced non-stationary effluent-concentration setpoint tracking with large tracking-error integrals (ITAEs) and offsets, whereas the RL agents tracked the setpoints with small ITAEs, no offsets and low variability, because their PID parameters were adapted online while those of the conventional controllers stayed constant.

Cast in the notation of Section 4, the action of a Class I agent at tuning episode $j$ is an increment of the gain vector, projected onto an admissible set $\mathcal{K}$ that preserves closed-loop stability, and a natural reward is the negative tracking-error integral of Eq. (13) over the episode:

$$\mathbf{K}_{j+1} = \Pi_{\mathcal{K}}\big(\mathbf{K}_j + \mathbf{a}_j\big), \qquad \mathbf{K} = \big[K_p,\, T_i\big]^{\mathsf{T}}, \qquad r_{j+1} = -\,\mathrm{ITAE}_j \tag{23}$$

The agent's decision interval is therefore the length of a tuning episode, many sampling intervals of the loop, and the projection $\Pi_{\mathcal{K}}$ is the means by which the gain limits a commissioning engineer would impose are made explicit.

The motivation for this class is well established in HVAC practice. Loop tuning is "currently somewhat of an art" performed empirically by trial and error: the proportional gain is tuned first and the integral gain then adjusted to eliminate offset, the derivative gain is usually not used at least partly because it complicates the tuning process, initial values are taken from rules of thumb, manufacturer recommendations or experience with similar applications, and the technician raises the gain until oscillation is detected before backing off to about one-half of that value [22]. The procedure, while not optimum, usually provides reasonable results, and its outcome depends on the experience of the technician; PID values may also perform differently from one manufacturer to another [22]. The gain-variation problem that motivates online adaptation is intrinsic: the gain of the controlled system can change with operating conditions even after the device has been linearised [9]. Because the derivative term is largely absent from HVAC tuning practice [22], the action space of an HVAC gain tuner reduces in practice largely to the proportional and integral terms, and the objective is to hold the loop at the favourable end of the dependence of accuracy and stability on the selected gains [22].

Class I is also the point at which RL must be distinguished from classical adaptive control, with which it shares the outer-loop position. Model reference adaptive control compares the plant's actual output with a desired reference model's output and adjusts control parameters to minimise the error, while self-tuning control continuously estimates the system's dynamics and adjusts parameters using identification methods such as recursive least squares or instrumental-variable methods [24]. RL is treated in the same family as an adaptive technique that learns control policies by trial and error, with agents interacting with the environment, receiving feedback through rewards or penalties, and updating their strategies to maximise cumulative reward [24]. The architectural difference is the source of the learning signal: a reference model in the classical case, a scalar reward derived from the loop's own measured response in the RL case. That difference has a consequence the cited study makes explicit — a comparison of the framework against PID gain scheduling is left to future work. From the plant's point of view, Class I is the least invasive of the four: it requires read access to the controlled variable and write access to the gain parameters of a controller that already exists, and no new actuation path.

### *5.1.3 Class II — RL as direct replacement of the regulatory controller*

In the second architecture the agent absorbs the feedback function and maps measurements directly to actuator demands, so that the PID block disappears and the reward must carry everything the error signal previously carried. The standard formulation casts energy-cost minimisation as a Markov game and solves the resulting HVAC control problem with multi-agent deep reinforcement learning [27], while other work learns an operating strategy for building HVAC systems directly from data by deep reinforcement learning [27], and deep reinforcement learning methods such as Deep Deterministic Policy Gradients are reported to balance energy consumption and occupant comfort in HVAC systems [28]. Direct control of an air-handling unit by a PPO agent with $CO_2$-constrained ventilation and economiser logic is likewise reported to give better temperature stability and energy efficiency than GA-tuned PID and on/off control within the modelled environment [95]. The action space is typically continuous: in an optimal energy management problem the continuous actions comprise the charge and discharge power of the storage systems and are envisaged to extend to the heat and cold power produced by the thermo-refrigerating heat pumps and chillers, electric-vehicle smart charging, domestic hot water tanks and public lighting, on the argument that continuous spaces give precise, fine-grained control whereas discretised spaces inevitably lose granularity and produce sub-optimal performance [29].

The instructive finding for architecture is that replacement is rarely total. Storage constraints in that system are traditionally handled by a low-level controller outside the agent, which receives the action taken by the high-level control agent — an RL agent, an MPC, or another controller — and adapts it to meet the system constraints; this classical constraint handling ensures the constraints are met but does not provide the RL agent with sufficient feedback during training, which motivated adding penalty components to the reward signal and thereby circumventing the sparse-reward problem [29]. The class therefore retains a residual constraint layer beneath the agent at the actuator interface, and relocating constraint enforcement from that layer into the reward changes the learning signal rather than removing the layer. The interface between agent and plant is correspondingly demanding: because the agent writes the actuator demand directly, the design of the state vector, action vector and reward function is what adapts a general algorithm to a specific plant, and the comfort requirement must be encoded explicitly — in the benchmark comparison of controllers within a digital twin, all three controllers, including the rule-based one, were constrained by thermal comfort bounds [20].

### *5.1.4 Class III — RL combined with MPC in a hybrid hierarchy*

The third architecture places learning inside or alongside an optimisation-based controller. Two coupling directions recur. In the first, the optimiser acts as teacher: the optimal control strategy is learned from an optimised operation dataset, and a scenario-reduction method that clusters similar operation scenarios dramatically reduces the number of optimisation calculations, significantly increasing the speed of generating those datasets while achieving energy savings [27]. In the second, the learner is trained through a model of the plant and deployed as the policy: a differentiable predictive control policy was trained end-to-end through a modularised sequence-to-sequence surrogate whose four submodules mirror first-principles energy balance — external heat transfer, internal heat gains, HVAC load injection and inter-zone conduction — each implemented as a neural encoder–decoder pair trained with physically consistent constraints, with phase-change transitions captured through data-driven hidden states without explicit enthalpy modelling; the policy minimised a multi-objective cost combining comfort within prescribed setpoint bounds, time-varying energy costs and peak power demand [30]. Execution was achieved by co-simulation with EnergyPlus, using Energy Management System callbacks to inject optimised air mass flow rate and supply temperature at four stages of the simulation pipeline — initialization of the heat balance, load prediction, HVAC manager update and final load correction [30].

The hierarchy into which such a controller fits is a generic one in process control: an upper structure separates scheduling, optimising and controlling problems into layers according to computational effort and operation horizon, with the upper optimisation layers minimising a cost function and sending the optimal schedule or operating points to the control layer, which seeks stable operation, reference tracking and disturbance rejection; the optimisation layers typically use real-time optimisation while the multivariable control layer uses MPC, and both run models continuously in a feedback loop so that model accuracy directly affects control, optimisation and hence process operation [31]. For intermittently operating concentrating solar collectors with a wide operational range and both slow and fast dynamics, real-time optimisation with nonlinear steady-state models combined with linear MPC using dynamic models presents limitations [31]. MPC itself is a family of control strategies directly dependent on physical models, whose use of models together with forecasted inputs allows system dynamics to be considered in control planning — a feature especially attractive for high-inertia systems such as buildings with high thermal mass [16].

What the learning agent inherits in this class is the optimiser's information requirement. An economic MPC formulation integrates a calibrated thermal model, short-term forecasts for weather and on-site generation, electricity price signals and constraints to coordinate HVAC and electrical energy storage, jointly minimising grid import costs and deviations from preferred setpoints while shifting load from high-price to low-price periods [32]. The model must predict building response to actuation and account for physical properties, weather predictions and occupancy predictions, all of which are highly heterogeneous between buildings, and the interface to the physical installations varies with the building automation degree actually installed [16]. The reported outcomes of this branch define the performance that a hybrid must preserve: an MPC framework based on a data-driven model with exogenous inputs reduced heating energy by around 12% with less than 0.5 °C impact on comfort [7]; a learning-based MPC deployed in real buildings achieved statistically significant savings of up to 1.5 MWh per day while maintaining thermal comfort, and MPC applied to an office HVAC system over a two-month deployment yielded approximately 10% savings relative to a baseline static configuration [18]; MPC is credited with anticipating future thermal disturbances, managing multivariable system interactions and operating within system constraints [18]. End-to-end architectures linking a data-driven MPC digital twin to existing supervisory control and data acquisition (SCADA) or building energy management systems, and MPC augmented with data-driven model-error compensation, are the practical forms this interface takes [7].

### *5.1.5 Class IV — RL for setpoint and thermostat scheduling at the supervisory level*

In the fourth architecture the agent never touches the regulatory law at all; it decides the reference that the loop tracks. Reinforcement learning and predictive control have been used to maximise HVAC setpoints [17], and multi-objective optimisation with NSGA-II has been applied to simultaneously optimise insulation levels and thermostat setpoints, producing notable improvements in comfort and energy efficiency across different Iranian climate zones [17]. The setpoint itself is a substantive decision variable rather than a fixed parameter: the conventional method sets a static 20 °C in winter and 24 °C in summer according to ASHRAE 55, whereas an adaptive setpoint is not fixed but changes proportionately to the monthly average outdoor temperature, and occupancy-driven thermostats based on adaptive comfort principles can significantly reduce energy consumption by 30–50% even in air-conditioned buildings [2]. The magnitude of what this layer controls is visible in simulation of scheduled versus continuous HVAC operation for phase-change-material envelopes: annual savings were 20.5% under continuous operation and 21.5% under scheduled operation, with summer savings rising from 47.5% in June to 51.2% in August for the single-layer configuration and from 70.2% in June to 73.2% in July for the double-layer configuration, because intermittent plant operation produces greater wall-temperature variations and thus better exploitation of latent storage [33]. Synchronising HVAC operation schedules with the thermal characteristics of phase-change materials, for instance through pre-cooling or thermal drift, can enhance latent storage advantages without sacrificing thermal comfort, although the ideal interaction between such envelopes and HVAC controls remains largely unexamined, particularly regarding energy flexibility and demand response [17].

In the MDP of Section 4 the Class IV action is the setpoint itself, bounded by the comfort band of Eq. (18) and issued at a supervisory interval that is a large multiple $N$ of the DDC sampling interval $\Delta t_s$:

$$a_t = T_{sp,t} \in \left[T_{low}, T_{up}\right], \qquad \Delta t_a = N\,\Delta t_s, \quad N \gg 1 \tag{24}$$

The hydronic study of Brandi et al. is an instance of this class, the agent writing the supply-water temperature setpoint that the terminal-unit loops then track [90].

The interface for this class is a setpoint write, and it can be introduced in stages. One digital twin framework is designed for phased integration — passive monitoring, then guided adjustments, then autonomous control — retrofit into existing systems gradually through building management system interfaces without replacing the underlying HVAC control hardware, and at the autonomous stage the twin actively sends optimised setpoints to BMS controllers, for example via BACnet commands, thereby taking on a high-level control role [19]. Where automation is restricted, the same output can be routed through people: building owners may opt to implement recommendations manually if they face policy or operational restrictions on automated control [28], and a human-centred digital twin energy management system couples HVAC and miscellaneous electrical load energy use with occupancy data in a building information modelling environment, with a cloud server performing the analysis and integration and the visualised information used as feedback to adjust the systems in real time [34]. The layered separation that makes such an interface possible is drawn explicitly in a three-layer building architecture whose control-hardware layer receives signals from sensors and issues commands to equipment while implementing low-level control strategies to keep actuators within acceptable conditions, and whose processing layer above it carries the computational elements for advanced control [35]. In SCADA terms this is the top of the hierarchy, where decisions and directions are made by the central computer and where production planning and collective strategies and scheduling are set [11]. An adaptive precedent for the outer tier exists in optimum-start programs, which start an HVAC system as late as possible in the morning while still reaching the comfort range by occupancy, using factors based on previous building response, system characteristics and current weather, and which monitor controller performance by comparing the actual and calculated time required and calculating new factors to improve it [8].

*5.1.6 Comparison across classes: plant, sensing and commissioning burden*

Read against one another, the four classes order consistently by what they demand of the plant. Class I requires no more than write access to the parameters of a controller that is already installed and commissioned, and its learning signal is derived from the same controlled variable the loop already measures. Class II requires full actuation authority and therefore a residual constraint layer at the actuator interface, together with an explicitly encoded comfort requirement in the reward [29]. Class III requires a calibrated thermal model, weather and on-site generation forecasts, price signals and an optimisation framework in addition to the loop itself [32], and the model must reproduce building response to actuation under heterogeneous physical properties, weather and occupancy [16]. Class IV requires only a writable setpoint terminal and a schedule, as the staged BMS-interface design shows [19].

Sensing is largely common to all four, but the marginal requirement differs. The underlying measurement set is the same in every case — humidity and temperature monitors, $CO_2$ and indoor air quality sensors, motion and occupancy sensors, light sensors measuring lux, and sensors for air and water pressure and flow — and it is these measurements that allow airflow and cooling output to be adjusted to real occupancy and heat gains, lighting to be dimmed or switched when daylight suffices, and ventilation rates to be modulated by demand-controlled ventilation [36]. What changes is the breadth of the state description the agent consumes: Class I uses the loop's own controlled variable; Class II depends on the full state vector designed for the agent; Class III adds exogenous weather, generation and tariff feeds [32]; and Class IV, where comfort is the objective, can require derived indices such as predicted mean vote computed from environmental and personal data by a dedicated user agent before the HVAC equipment executes commands [35]. Across all four, the data layer is where the technical complexity concentrates, since it involves the collection, integration and transmission of data across subsystems, with the primary challenges being high-quality data — particularly from building information modelling and sensors — and the integration of existing building management, geographic information and weather systems [37]; deployments in building operation and maintenance additionally face data quality and reliability, data management and privacy, sensor limitations, and high implementation cost with a lack of skilled personnel to maintain the system [38].

Commissioning effort follows the same ordering but with a shift in kind. Class I displaces a task that is already labour-intensive and person-dependent, since tuning proceeds empirically by trial and error from rules-of-thumb or manufacturer values, its precision scales with technician experience, and the resulting procedure, while not optimum, is usually judged reasonable [22]. Class II shifts the effort into reward and state design and into training under the uncertainty the plant will actually encounter, since the time-varying uncertainty must be incorporated into agent training for the adaptation to hold. Class III shifts it into modelling and interface engineering, which the MPC literature identifies as the principal practical obstacles: an interface towards the sensors and actuators whose configuration, installation and deployment depend on the physical properties of the building, and a model able to predict building response to actuation [16]. Class IV shifts it into schedule and setpoint curation and, where automation is restricted, into human review of recommendations [28]. Verification is the least settled part of this comparison: where operational modes are implemented sequentially over different periods rather than concurrently, seasonal weather variation and occupancy change confound direct comparison, and the authors of one such study segment their data into comparable condition groups and construct a balanced dataset precisely because performance differences cannot be attributed to the control strategy with confidence [19]; a digital twin can partially offset this by disaggregating whole-unit consumption to room level, spatial granularity that a single installed energy meter cannot provide [20]. Reviews of the field correspondingly screen studies for experimental validation and measurable indicators such as energy savings, comfort indices or forecasting accuracy, and exclude conceptual work without such quantification [39].

The unifying observation is that the physical loop is the same object in all four architectures, and only the terminal through which the agent intervenes changes. The chain from sensor through controller and controlled device to controlled variable persists in each case [9], physically embodied in DDC unitary controllers preprogrammed for particular equipment applications [12] and in field controllers that generate demand signals from preprogrammed algorithms [11]. Class I writes to the parameters of the controller; Class II writes to the actuator demand; Class III writes the reference trajectory that the loop is asked to follow, computed upstream in a hierarchy whose upper layers allocate schedules and operating points [31]; Class IV writes the setpoint that defines that reference [19]. The classes are therefore not competing philosophies of control but nested points of attachment on one plant, and hybrid implementations can occupy more than one at once — as in a policy trained through a differentiable surrogate but executed as actuator commands injected into the simulation loop [30], which uses the interface of Class II while resting on the model-based information structure of Class III.

# 6. Comparative Synthesis of Control Strategies

## *6.1 Comparative Synthesis of Control Strategies*

The four families compared in this section differ less in the variable they regulate than in where the burden of knowledge about the plant resides: in a fixed schedule, in three hand-tuned gains, in a learned policy, or in an explicit model or constraint set retained alongside a data-driven component. **Table** 2 consolidates the comparison across eight criteria, drawing on the evidence reviewed above. The hybrid column is used here in a deliberately broad sense to denote architectures that keep an explicit model, physical constraint or rule base in the loop alongside a data-driven or optimisation component — RL-tuned PID, physics-informed or safety-projected learning, and model-based predictive control (MPC) — because these approaches share the property that neither the model nor the data-driven element alone specifies the control action [40]. Several of the latency and constraint-satisfaction figures in **Table** 2 originate in power-system control benchmarks; they are reported here as evidence of how each architecture behaves under formal evaluation rather than as HVAC-specific performance values.

**Table** 2. Consolidated comparison of thermostatic/on–off, PID, RL and hybrid control architectures across eight criteria.

| Criterion | Thermostatic / on–off and fixed-schedule rule-based | PID / PI | RL / DRL | Hybrid (MPC, RL-tuned PID, physics-informed or safety-projected learning) |
|---|---|---|---|---|
| **Data and sample efficiency** | No training corpus; operation follows static schedules and predefined set-points without real-time sensor feedback [19] | No training corpus; tuning requires a steady-state plant response and representative operating conditions [9] | Learns by interaction with the environment under a numerical reward, and its computational and memory requirements grow exponentially with the number of state and action variables [29]; deep Q-learning controllers reduced energy costs by up to 45% while holding comfort within 1.2 °C of set-point [30] | Physics terms allow physical relationships to be learned from measurements directly, reducing reliance on pre-solved labels [41]; physics-informed formulations are reported to perform on small datasets [42] |

| Criterion | Thermostatic / on–off and fixed-schedule rule-based | PID / PI | RL / DRL | Hybrid (MPC, RL-tuned PID, physics-informed or safety-projected learning) |
| --- | --- | --- | --- | --- |
| **Stability and performance guarantees** | Two-position action causes a minimum of 2 °F (1.1 °C) room-temperature swing as the thermostat toggles the air valve [13]; rigid logic produces part-load inefficiency and overcooling against conservative safety margins [19] | PI holds variation below 2 °F (1.1 °C) depending on the integral setting; PID holds below 0.5 °F (0.28 °C) or better depending on the derivative setting [13]; high plant gain tends to produce instability, and a controller tuned at high gain may perform sluggishly at low gain [9] | Pure learning-based methods suffer constraint violations and lack optimality guarantees, making them unsuitable for safety-critical applications [43]; reward-based soft constraint enforcement provides no theoretical hard safety guarantee at every decision step [44] | A differentiable power-flow projection layer with a safety-net solver fallback guarantees hard feasibility [45]; clamping and Sigmoid projection embed operating limits directly in the computational graph, giving deterministic constraint satisfaction [46] |
| **Interpretability** | Control logic is deterministic and derived from historical operational patterns [19]; self-optimising control holds a constant set-point for a selected variable that indirectly yields near-optimal operation [47] | The control law is fully specified by proportional, integral and derivative gains whose effect on step response is directly observable [22] | Limited interpretability of black-box decisions is one of three persistent limitations of learned controllers [30]; opacity of model decision mechanics produces user distrust [48]; ML models are the least interpretable class, requiring SHAP or LIME to expose input influence [49] | Grey-box formulations whose coefficients carry physical meaning are more interpretable than ML models, while non-ML black-box models fall between the two [49]; a PWARX model identifies discrete thermal states across phase transitions whereas LSTM lacks explicit phase-change detection [50] |
| **Commissioning effort** | Low: schedules and fixed set-points are set once and serve as a baseline against which other modes are evaluated [19] | Loop tuning is "somewhat of an art" performed empirically by trial and error, adjusting gain upward until oscillation and backing off to about one-half of that value [22]; autotuning substantially reduces commissioning time but requires steady state, representative conditions and a deliberate order of loop tuning [9] | Requires substantial policy training and online updates, which raises computational burden relative to single-forward-pass supervised prediction [51]; structured, scalable solutions that automate data collection, model calibration and control-signal deployment in a live context are lacking [32] | Extends commissioning into continuous model updating, so that predicted demand stays close to real operation [52]; digital-twin configuration differs by intended service — ongoing commissioning, performance optimisation or retrofit analysis [37] |

| Criterion | Thermostatic / on–off and fixed-schedule rule-based | PID / PI | RL / DRL | Hybrid (MPC, RL-tuned PID, physics-informed or safety-projected learning) |
|---|---|---|---|---|
| **Sensor requirements** | One sensor per zone suffices under the ASHRAE 90.1-2022 HVAC-zone definition, a space or group of spaces controllable from a single thermostat or temperature sensor [10] | Local feedback on the controlled variable: the test-house installation used A-class PT100 temperature sensors and an electromagnetic flow meter driving a PID-controlled heater through a solid-state relay [25] | Increased sensor deployment widens the applicability of black-box modelling [7]; a digital-twin fault-detection framework combined BIM, sensor data and occupant feedback in real time with a 5 min response time [1] | Grey-box models exploit both real-time sensor data and the building's intrinsic characteristics [7]; data-quality control predicts faulty sensors from fault-free sensors in the same time instant, which presupposes redundant sensing [53] |
| **Computational load** | Negligible; control sequences reside in standalone zone-level controllers and are selected and set up through a PC or portable operator terminal [8] | Parameters remain fixed between tuning sessions, and tuning is an occasional, engineer-initiated action rather than a continuous computation [9] | Training dominates: a physics-informed network required 0.66 s per epoch against 0.17 s for a vanilla network [54]; ML models show considerably higher prediction time than non-ML models, with RF the highest [49]. Once trained, inference runs as a single forward pass at millisecond latency [55]; a physics-informed GNN achieved 0.05–0.38 ms inference on CPU, over 4,000× faster than conventional solvers [45] | The physical penalty term adds computation, making PINN4PF more expensive than black-box networks and exposing an accuracy–cost trade-off [56]; a hybrid trapezoidal solver in which one generator is replaced by a PINN permits larger time steps and reduces the number of global integrating steps [57] |

| Criterion | Thermostatic / on–off and fixed-schedule rule-based | PID / PI | RL / DRL | Hybrid (MPC, RL-tuned PID, physics-informed or safety-projected learning) |
|---|---|---|---|---|
| **Safety** | Conservative safety margins lead to unnecessary equipment operation or overcooling under part-load and mild-weather conditions [19]; two-position containment control methods either ignore, assume or manually fix the pressure offset [12] | A high system gain tends to produce control instability, so controllers must be tuned at high gain; an autotuner cannot compensate for variations in system gain between sessions [9] | Soft constraint enforcement through reward penalties and early termination does not provide theoretical hard safety guarantees at every decision step [44]; data-driven models may produce predictions that violate fundamental power-flow equations or operational constraints [58] | Cascaded differential-flow/differential-pressure control combines the pressure accuracy of DP and the stability of DF [12]; combining machine learning with expert rules enables prompt detection and resolution of issues [1]; a physics-informed controller held maximum voltage deviation below 0.02 pu against 0.06 pu and 0.04 pu for droop and VSM, with recovery in about 1.5 s against 3 s and 2.5 s [59] |
| **Retrofit feasibility** | The incumbent mode in legacy plants, modelled by enforcing deterministic logic derived from historical operational patterns [19]; on/off pressure switches can be replaced by PID-type pressure controls and the burner derated [15] | Standalone zone-level controllers host several resident control sequences for unitary equipment, selected and set up on site [8]; a low-cost Raspberry Pi-based control and data-acquisition system implemented PID control of a test house [25] | Few studies demonstrate full-scale deployment in real buildings using commercial hardware and infrastructure [32] | Digital twins depend heavily on accurate and complete BIM data that is rarely available in operational buildings [28]; solutions able to integrate BIM yet not solely dependent on it are therefore of significant interest for the existing stock [37] |

### *6.1.1 Data and Sample Efficiency*

The first criterion separates the families most sharply, because two of them consume no training corpus at all. Fixed-schedule control operates from static schedules and predefined set-points without real-time feedback [19], and PID tuning extracts parameters from an observed plant response under steady-state conditions rather than from historical data [9]. Reinforcement learning, by contrast, learns by interacting with the environment and refining its strategy through trial and error against a numerical reward [29], which places the burden squarely on the quantity and coverage of experience. The historical obstacle was dimensionality — computational and memory requirements grow exponentially with the number of state and action variables [29] — and the reported field performance of deep Q-learning controllers, including energy-cost reductions of up to 45% at comfort held within 1.2 °C of set-point, ice-storage scheduling savings of up to 10.8% monthly, and reinforcement-clustering sequencing that lowered consumption by over 30%, is achieved on the far side of that burden [30].

The learning-based studies in power-system control give a sense of the sample regime this implies in practice: contemporary training sets are on the order of $10^4$–$10^5$ samples [60], and one power-flow study built 16,000 sample points per test system and split them into supervised, unsupervised and test partitions [61]. Returns diminish with volume rather than scaling indefinitely: in a hybrid physics-guided scheme the generation-cost deviation fell from 3.5% at 10,000 training points to a stable 0.4% at one million, with the network-loss error reaching 0.3% against 1.6% for a CGAN-plus-stochastic-OPF baseline and 3.3% for DNN regression [62]. The corresponding caveat is that machine learning often requires high-quality training data and remains prone to overfitting in complex building environments [28]. Hybrid formulations intervene precisely at this point: embedding physical relationships allows the model to learn from measurements directly and reduces reliance on pre-solved labels whose global optimality is difficult to certify [41], and physics-informed methods are reported to remain workable on small datasets [42].

**Figure** 6 collects the headline savings reported by the studies reviewed here. The spread within the RL group alone, from 5–12 % of heating energy [90] to 20–70 % of energy cost [89], is as large as the spread between groups, consistent with the observation that results are expressed against different baselines, metrics and validation regimes; the figure is therefore a map of reported claims rather than a ranking of architectures.

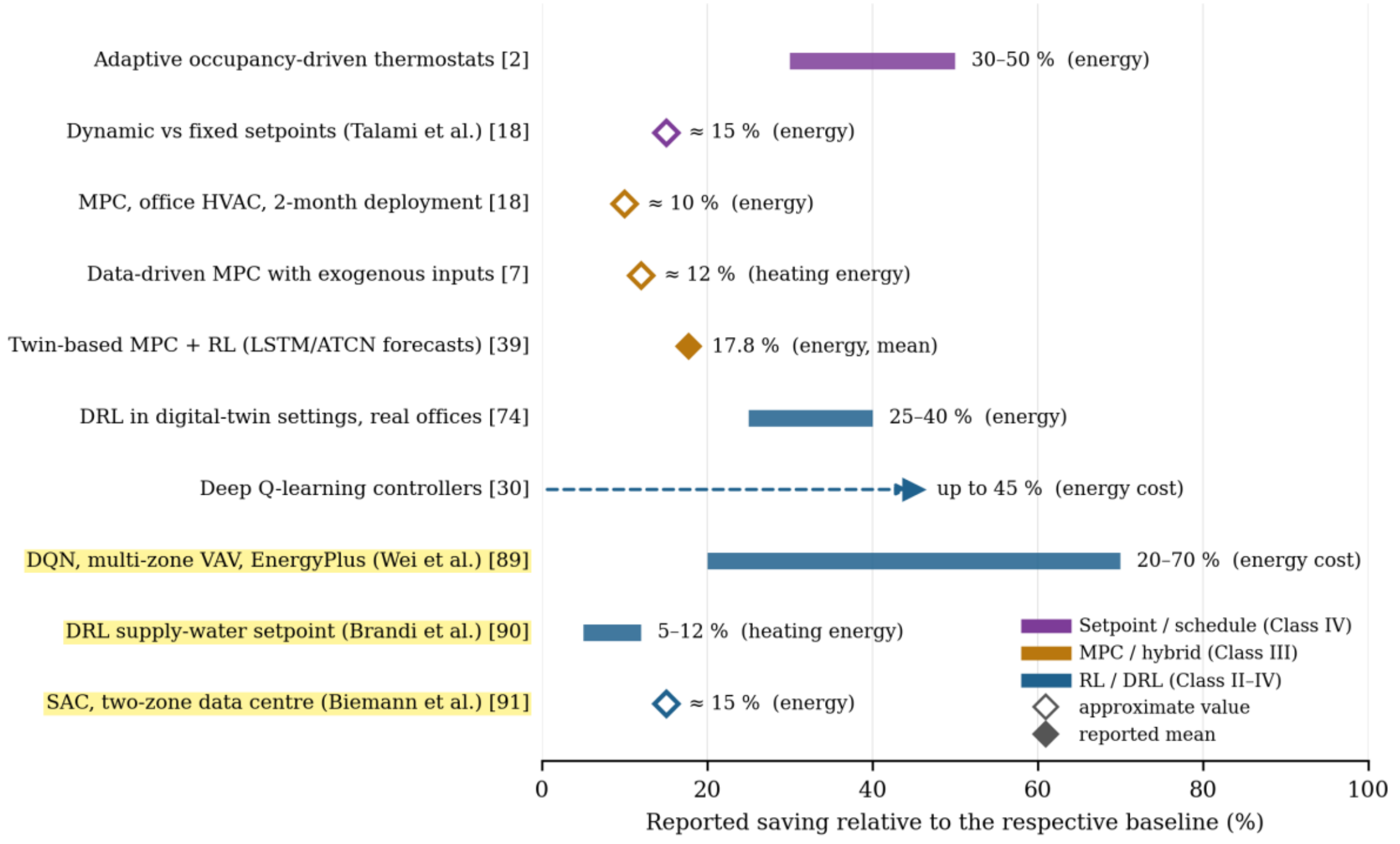

**Figure** 6. Headline savings reported by the reviewed studies, grouped by control architecture [2, 7, 18, 30, 39, 74] and [89]–[91]. Bars denote reported ranges, diamonds single values, the arrow an upper bound. Baselines (rule-based, static schedule or fixed setpoint), metrics (energy, energy cost or heating energy) and validation

regimes (simulation or field) differ between studies, so values are not directly comparable. Chart prepared by the authors.

## 7. Physics-Informed and Digital-Twin-Enabled Control

### *7.1 Physics-Informed and Digital-Twin-Enabled Control*

The architectures set out in Section 5 differ most sharply in what they assume about the plant: a static schedule assumes nothing, a predictive controller assumes a calibrated dynamical model, and a learned policy assumes an environment in which many candidate actions can be evaluated without cost to occupants. The two technologies examined here supply those assumptions. Physics-informed learning restricts the hypothesis space of a learned model to functions consistent with the governing conservation relations, while the digital twin supplies the environment in which a candidate controller can be exercised, benchmarked and falsified before it is allowed to actuate. Their joint function, argued below, is to move the verification of an HVAC controller off the live plant and onto a model that can be interrogated repeatedly.

#### *7.1.1 Conservation relations as a constraint on the learning problem*

A physics-informed neural network (PINN) improves the physical consistency of neural predictions by embedding governing equations or residual constraints into the training loss [63]. The formulation introduced by Karniadakis and colleagues approximates partial differential equations by integrating their governing equations together with their initial and boundary conditions into the network's loss function, exploiting automatic differentiation in frameworks such as TensorFlow and PyTorch to compute the required derivatives [64]. In contrast to conventional supervised learning, such networks do not require labels, since training data can be generated on the fly and self-supervision is supplied by the known equations; because the input space need not be discretised, the curse of dimensionality is avoided, and systems of differential–algebraic equations can be solved in a fraction of the time taken by traditional methods, with state variables obtainable at any instant $t_1$ without integration from $t_0$ [64]. In its most common implementation the objective is a data term augmented by weighted residual penalties, $L_{PINN} = L_{data} + L_{PI}$ with $L_{PI} = \sum_{i=1}^{m} \beta_i \times PI_i$, so that the physical equations participate in training through a penalty on their iteration residuals [58].

The motivation is the pair of failure modes that a purely physical or a purely empirical model each exhibits. Purely physics-based models depend completely on first principles and mathematical formulations, are usually computationally expensive, and may require substantial simplifications that discard important system detail, whereas purely data-driven models recognise nonlinear relationships directly from data but disclose little about the underlying physics, usually require large amounts of data, and may fail to generalise beyond the range of the training set [26]. In buildings the first difficulty is acute: as HVAC systems become more complex, nonlinear and large-scale, physics-based models become harder to develop, high-order models are computationally expensive, reducing model order increases prediction error, and long computation times make complex models difficult to implement in real time, while the resulting models are deterministic and rest on multiple assumptions and simplifications [5]. The black-box alternative is no more satisfactory on its own, since data-driven models built straight from data cannot be easily comprehended or interpreted to explain how variables combine to produce a prediction [5]. Physics-informed machine learning therefore occupies a systematic middle ground between purely model-free regression and full online optimisation, embedding physical structure, residual penalties or constraint awareness into the learning objective [55].

For thermal systems the natural constraint is the energy balance. A modularised physics-consistent neural network for zone control was grounded in the first law of thermodynamics, with each zone governed by $C_z\, dT_z(t)/dt = \dot{Q}_{ext}(t) + \dot{Q}_{int}(t) + \dot{Q}_{hvac}(t) + \dot{Q}_{iz}(t)$, where $C_z$ is the zone's effective thermal capacitance and the right-hand terms are the heat flows from the external envelope, internal gains, HVAC system and interzone interaction respectively [30]. Comparable embeddings have been reported for multi-zone electric water heaters, where multiple-input–multiple-output and hybrid multiple-input–single-output PINNs enhanced with physics-derived features and recursive training outperformed traditional differential models in accuracy, interpretability and computational efficiency on real-world data [54], and PINNs have been applied to control-oriented modelling of existing buildings, transient heat conduction and temperature-field inversion of heat-source systems [65]. Where the governing differential heat equation is not embedded, physical consistency remains partial: an LSTM model of a mortar wall containing 15 wt% microencapsulated phase change material reached $R^2 = 0.99$ yet lacked the intrinsic capability to offer detailed physical interpretation, and the authors identify a PINN incorporating the fundamental heat transfer equation with convection effects as the route to improved generalisation and interpretability under varying hygrothermal conditions [50]. In the load-forecasting setting, physics-informed adaptations of LSTM and GRU models improved prediction accuracy and interpretability relative to their unconstrained counterparts in a smart-home application [35].

Two qualifications matter for control. First, physics can be injected either through the network architecture or through the loss function, the former using topology-aware structures and non-learnable physics modules that penalise violations of conservation and imbalance relations, the latter penalising equation residuals or constraint violations [66]. Second, a soft penalty is not a guarantee: penalising residuals during training provides no hard feasibility guarantee, and a more principled route implements the governing equations as differentiable constraint layers that project every prediction onto the feasible manifold, reducing residuals well below operational tolerances with a solver fallback for hard feasibility [45]. Physics-informed formulations have raised feasibility rates while leaving the satisfaction of constraints during inference a fundamental open challenge [43], and a penalty formulation remains different from a hard projection or an exact solve [67]. Where physical knowledge enters only through the training loss, the architecture may still fail to exploit the structural couplings that distinguish operating conditions [63]. It is for this reason that physics-informed machine learning for safety-critical applications is increasingly framed around transparency, interpretability, explainability and explicit correctness verification rather than accuracy alone [68].

### *7.1.2 Physics-consistent surrogates as the plant model inside a controller*

A surrogate that respects the governing relations serves simultaneously as a model and as a computational accelerator. Once trained, PINNs used as an alternative to classical numerical integration approximate component dynamics more accurately over longer time steps and speed up simulation by substantially increasing the step size used [57]; reported gains include determination of rotor angle and frequency up to 87 times faster than conventional methods [69] and acceleration of electromagnetic transient simulations or reduced-order models by more than 100 times [70]. The data economy is equally relevant to buildings, since physics-informed networks require substantially less training data and can adopt simpler structures while achieving high accuracy [69], and an unsupervised physics-informed formulation minimising power-mismatch residuals has been shown to eliminate the need for high-quality labelled data while retaining generalisation, with gradients supplied by automatic differentiation for downstream optimisation [61]. Crucially for the architectures of Section 5, PINN surrogates used inside a reinforcement-learning loop enabled 50 % faster policy training at comparable performance while reducing reliance on costly simulator samples [54]; physics-informed and hybrid surrogates retain the speed of learned models while improving physical consistency, which is precisely what matters when purely data-driven models are exposed to unseen operating conditions or strong nonlinearities [71].

Within thermal plant specifically, the surrogate must be fast enough for the control interval it serves. A neuro-fuzzy ANFIS model of an absorption chiller ran three days of operation in 8.9 s at a simulation time step of 0.68 ms against a 20 s sampling time, and was therefore judged fast enough for model predictive control, achieving a worst-case mean error of 0.27 °C and a worst-case standard deviation of 3.58 °C at the 95 % confidence interval, embedding the inaccessible proprietary chiller controls and remaining re-trainable to cope with plant ageing or process modification [72]. An ANN prediction model of the output temperature of a ten-loop flat-plate solar collector field, connected to the physical plant through OPC-UA so that the twin receives sensor data continuously, supports operator training, online fault detection and system analysis [73]. The trade-off between transparency and accuracy recurs here: a piecewise ARMAX model identified discrete thermal states associated with phase transitions and explained them, whereas the LSTM counterpart forecast accurately without offering detailed physical understanding [50]. Deployment constraints close the loop on model choice, since embedding a surrogate in constrained hardware motivates physics-aware pruning, quantisation and model distillation that preserve accuracy while reducing memory and latency [59].

### *7.1.3 Digital twins as training, benchmarking and verification environments*

A digital twin is a virtual representation of a physical asset enabled through data and simulators for real-time prediction, optimisation, monitoring, controlling and improved decision making [31]; in buildings it is a dynamic virtual representation reflecting real-time condition through sustained data exchange, combining Building Information Modelling (BIM) with Internet of Things sensor networks, high-fidelity simulation engines and advanced analytics for predictive monitoring, simulation and optimisation [74]. Its core enabling technologies are modelling, simulation, verification and validation [31], which is why the twin functions as an assurance instrument and not merely a visualisation. The review literature in this area has been fragmented across building architectures, motivating an explicit bridge between twin architectures and the physics and control of thermal energy systems [23], and full-scale twins of thermal networks remain scarce, with only one study reporting a full-scale twin of a district heating and cooling network [31].

As a benchmarking environment, a calibrated twin provides the controlled, instrumented and repeatable conditions that a live building cannot. A pipeline for creating digital twins of building energy systems has been released as an open test environment for controller benchmarking, calibrated on an extensive dataset spanning a wide variety of data at fine temporal resolutions, and distinguished from existing test environments by its comprehensive list of controllable variables, the temporal resolution of its measurements and its real-time capabilities, with a case study demonstrating its use for benchmarking building automation and control system performance [20]. Comparable staging is explicit in hotel HVAC twins, where a Fixed Mode reproducing conventional rule-based operation with static schedules and predefined setpoints and no predictive or adaptive behaviour serves as the reference point against which the predictive and prescriptive/autonomous modes are evaluated [19]. The same logic underpins twin-based predictive control: a digital twin framework for MPC-based building automation combines a generic zone control application, a genetic-algorithm optimisation framework, a parametrised grey-box zone model and a data layer providing uniform access to sensors, actuators and weather forecasts alongside an occupancy prediction algorithm [16], an instance of which controlled a university classroom in Odense and held comfort levels comparable to commercial control strategies while permitting energy-saving strategies, improving on default rule-based operation [37].

As a training environment, the twin supplies the episodes that a learned controller requires. Deep reinforcement learning controllers deployed in digital twin settings have shown 25–40 % energy savings in HVAC systems in real office buildings in Switzerland and Norway [74], and a wireless-sensor-driven twin combining LSTM and ATCN forecasting with model predictive control and reinforcement learning reduced energy consumption by an average of 17.8 % relative to baseline rule-based controls while the Predicted Mean Vote index remained within the ASHRAE comfort range during 92 % of occupied hours [39]. Where a system is still under construction and no historical data exist, a Modelica twin can serve mainly as a testbed for the algorithms, to be replaced by a twin derived from real-world data once sufficient operational data become available [29]; the same study selected Deep Deterministic Policy Gradient on grounds of sample efficiency and generalisation, citing evidence that among the deep reinforcement learning agents tested against an MPC controller and rule-based controllers for a residential heat pump, only DDPG outperformed the baseline in all heating scenarios considered [29]. Twins also carry diagnostic duties that condition safe operation: a twin-enabled fault detection and diagnosis process identifies abnormalities that would otherwise produce poor indoor air quality, thermal discomfort and low productivity, using semantic knowledge tags and Bag-of-Words feature selection because excess sensory data degrades diagnostic performance and causes overfitting between faults and irrelevant dimensions, and updating the semantic model so that the twin evolves [21]; this matters economically because maintenance represents 65 % of total facilities management cost and HVAC maintenance is a particularly critical component of it [1]. Predictive maintenance models coupling Bayesian networks with rule-based fault detection have reported diagnostic accuracies of 91–97 % in non-residential Norwegian buildings [74]. Twin environments have likewise been used to simulate real-time environmental conditions and personal comfort responses, permitting scenario-based comparison of room-level against personalised thermal adjustments and identification of comfort zones that minimise energy use [75].

The instrument has limits that a control-engineering readership should register. Twins have attracted critique for opaque objectives towards optimisation and efficiency [4], and in energy-audit practice they are frequently implemented as passive visualisations rather than active diagnostic tools, with performance classification decoupled from behavioural data such as occupancy patterns or HVAC schedules [28].

*7.1.4 The simulation-to-plant gap and the role of calibration*

The distance between a twin and its building is empirically large when behaviour is omitted rather than modelled: ignoring occupant behaviour can produce a performance gap of up to 300 % between predicted and actual energy use, and annual household electricity consumption in southeast China has been reported to range from 562 kWh to 5320 kWh — a nearly tenfold difference largely driven by occupancy schedules, appliance usage and window-opening behaviour [28]. Even where comfort equivalence is demonstrated in a twin-in-the-loop MPC experiment, quantifying energy savings precisely remains challenging because of complex factors such as the building's ventilation system and weather fluctuations [37]. Calibration is therefore the operative discipline, not a preliminary step. Twins are calibrated against extensive datasets at fine temporal resolution [20]; a semantic twin tailored to a building's type of use can support live data acquisition, recursive model calibration, short-term forecasting and closed-loop control through commercially available interfaces, with a fully automated online routine continuously updating model parameters from a rolling window of recent operational data so that the model remains adaptable to evolving conditions [32]. The same principle appears in commissioning-oriented workflows, where the simulation model is continuously updated as conditions change so that predicted energy demand is kept in close range of real operation [52]. Automatic updating itself introduces a data-quality problem, in that captured values must not originate from measurement or provoked errors; a methodology combining a neuro-fuzzy fault detector with a recurrent network predicting error magnitude has been applied to the twin of a real solar cooling plant for exactly this purpose [53]. Model currency must extend to plant ageing, which is why re-trainability is claimed as a property of control-oriented twin models [72].

With a calibrated twin in place, the three objects that a commissioning engineer actually adjusts can be evaluated off-plant. Gain tuning is the first. Adaptive and self-tuning controllers offer potential benefits in reduced commissioning time, increased actuator life and improved environmental control, yet have not found wide application in the HVAC industry, and their operation may be adversely affected if tuning is attempted in the presence of plant failure or periodic disturbances; an adaptive controller that continuously seeks to optimise its operation may also compensate for plant degradation and thereby delay needed maintenance action [9]. The underlying difficulty is structural: proportional control logic assumes linear processes, whereas most real processes are nonlinear, so PID logic may be very difficult to set up to maintain zero error under all conditions, and dynamic self-tuning is proposed as the mitigation, with the additional benefit of reduced commissioning time [22]. Pattern-recognition adaptive control developed for plants modelled as first-order plus dead time, which includes most HVAC systems, is claimed to be easy to use and to provide near-optimal control and has been used successfully in a number of HVAC situations [9], and conventional tuning itself divides into rule-based approaches such as Ziegler–Nichols and Cohen–Coon and model-based approaches such as Internal Model Control, with the nonlinear and time-varying character of practical systems requiring frequent re-tuning [3]. Repeated, disturbance-controlled evaluation of candidate gains is exactly what a fault-free twin affords and a live plant does not.

Setpoint schedules are the second object. Where short-horizon forecasting of roughly one hour ahead is embedded directly into supervisory setpoint decisions for cooling-water temperature and flow, operation becomes proactive rather than reactive and verified energy savings are obtained relative to static control [18]. Schedules interact with storage in ways that only simulation can expose: prior work has presumed standard thermostat settings, for instance 24 °C for cooling and 20 °C for heating, irrespective of the phase change material's characteristics, which may hide the full potential of latent thermal storage, whereas synchronising HVAC operation with the thermal characteristics of phase change materials through pre-cooling or thermal drift can enhance the latent storage benefit without sacrificing comfort, and multi-objective optimisation of phase-change envelopes has been reported to save up to 20 % of annual HVAC demand across climate zones [17]. Weight selection within a single aggregated cost function is a related tuning burden, since weights are adjusted until the output makes sense, changing priorities requires re-tuning, output is never guaranteed stable outside the test set, and a cost function tuned to one building cannot be transferred to another with a different underlying model [16]; a twin is the only place where such re-tuning can be rehearsed against a building-specific model.

Learned policies are the third. A control optimisation framework that learns the nonlinear thermodynamics of phase-change-enhanced buildings through modularised physics-aware neural surrogates validates its control policies in a high-fidelity, regulation-compliant simulation environment before targeting deployment in embedded controllers and smart-grid platforms [30], and cleanroom deployments link an automatically calibrated physics-based twin to deployment-ready predictive control, embedding roughly one-hour-ahead energy forecasting in supervisory setpoints and demonstrating verified savings relative to static control, with robustness shown across multiple climate zones and transferability claimed for other mission-critical, high-density facilities such as data centres [18]. Twin-based evaluation also closes the loop between comfort modelling and control, since scenario-based energy evaluations within the twin can compare room-level against personalised adjustments and identify the comfort zones that minimise energy use while maintaining thermal satisfaction [75].

*7.1.5 Which architectures need a twin, and which do not*

Reading the architectures of Section 5 against these requirements produces a clear ordering of model dependence. Static rule-based and fixed-schedule strategies sit at one end: they operate on static schedules and predefined temperature setpoints without consideration of real-time feedback from environmental sensors or internal loads, assume no predictive or adaptive capability, and are modelled inside a twin by enforcing deterministic control logic derived from historical operational patterns [19]. Such strategies require no twin and can be tuned and diagnosed against the plant itself, at the price of the inefficiencies that arise under part-load conditions or mild weather, when equipment runs unnecessarily or spaces are overcooled to maintain conservative safety margins [19].

Self-tuning PID occupies an intermediate position. Indirect adaptive methods employ a model of the process that is continually updated by comparison with the behaviour of the real plant and is then used to calculate updated control parameters, and this method may be combined with neural networks [9]. The model required is local and low-order — a first-order plus dead-time representation suffices for pattern-recognition adaptive control across most HVAC systems [9] — so this class can be commissioned on the plant, subject to the caveat that tuning in the presence of plant failure or periodic disturbance degrades operation [9].

Predictive architectures sit at the opposite end, because the model is constitutive rather than auxiliary. Economic model predictive control integrates a calibrated thermal model, short-term forecasts for weather and on-site generation, electricity price signals and constraints coordinating HVAC with electrical energy storage, jointly minimising grid import cost and deviation from preferred setpoints [32]; twin-based MPC frameworks likewise depend on a parametrised grey-box zone model together with occupancy prediction and a uniform data layer [16]. The choice among physics-based (white-box), data-driven (black-box) and hybrid (grey-box) dynamic thermal models is therefore not stylistic for this class but determines whether the controller can be formulated at all, and each approach begins with the definition of boundary conditions, for which the building's thermal zones must first be determined.

Model-free reinforcement learning lies in between in a specific sense: its formulation contains no explicit plant model, yet the evidence assembled here trains and evaluates it inside a twin — deep reinforcement learning controllers have been assessed in digital twin settings before claims of 25–40 % HVAC savings in occupied buildings [74], a twin has served as the algorithm testbed where real data were unavailable [29], and a surrogate built from physics-informed learning has been used inside the loop to halve policy training time while reducing dependence on costly simulator samples [54]. Architectures in which a learned agent tunes PID gains or adjusts setpoints inherit this requirement in full, since they need repeatable episodes over which gains and setpoints can be varied while the consequences for comfort and energy are measured [39]. Architectures built on physics-consistent differentiable surrogates require both ingredients simultaneously, the surrogate supplying the gradients and the regulation-compliant testbed supplying the verification [30]. Across all of these, the twin's value is bounded by the fidelity of its controllable-variable list, the resolution of its measurements and its real-time capability [20], and by the honesty of its constraint handling:

where feasibility is enforced by penalty rather than projection, a policy that appears admissible in the twin may still violate operating limits at the point of inference [45].

## 8. Open Challenges and Research Agenda

### *8.1 Open Challenges and Research Agenda*

Across the architectures reviewed here, the obstacles that recur are as much evidentiary and institutional as algorithmic: few learned control frameworks have been demonstrated at full scale in real buildings using commercial hardware [32], continuous real-building operation beyond 12 months is reported in only seven (14%) of the surveyed studies and in none beyond 24 months [74], and the hard guarantees that safety-critical operation presupposes are typically absent from reward-based formulations [44]. The five directions set out below are ordered from those that block deployment today to those that require longer-term methodological work.

#### *8.1.1 Embedding learned layers within existing building-management hierarchies*

The first obstacle is architectural rather than algorithmic. Existing frameworks are typically tested in simulation, and although digital representations have shown value for monitoring and diagnostics, their integration with predictive control remains rare in practice, with a lack of structured, scalable solutions that automate data collection, model calibration and control-signal deployment within a live operational context [32]. A workable migration path has, however, been demonstrated: the twin begins as a passive monitor, proceeds to guided adjustments, and finally sends optimised setpoints to BMS controllers—for example as BACnet commands—assuming a high-level control role while the underlying HVAC control hardware is retained, so that the solution can be retrofitted gradually [19]. Comparable couplings link data-driven MPC digital twins to SCADA or building energy management systems [7].

The lower loops with which such a layer must coexist are not static either. DDC systems already self-commission, retuning parameters as conditions change—altering PID loop parameters when a large air-handling unit moves from heating-coil control in winter to cooling-coil control in summer is the textbook example [22]—and optimum-start controllers learn the behaviour of building and plant and adjust their settings accordingly, an instance of adaptive control [9]. Networked digital controls likewise share points such as outdoor-air temperature among controllers and align chiller, pumping and fan strategies with system demand [8]. A learned supervisory layer therefore has to negotiate setpoint authority with loops that are themselves self-tuning, and existing experimental studies often rely on fixed configurations and fail to show how control frameworks adapt to varying levels of system integration [32].

Semantic structuring offers one concrete research route. Knowledge tags embedded in semantic models decompose complex systems into subsystems, making analysis computationally efficient through a divide-and-conquer strategy that supports localised decision-making [21], and an ontology-based data model has been proposed to establish standardised integration across heterogeneous sensors and application systems [76].

Standardised schemas such as IFC and BACnet, combined with modular APIs, are advanced as the substrate for interoperable audit platforms [28]. Interoperability, data harmonisation and scalability remain the recognised deployment challenges of this integration [23], and building digital twins have neither a standardised architecture nor a fixed set of services, their features varying with intended function [37]. Specifying interface contracts between learned and conventional layers—authority limits, fallback rules and hand-over conditions—is consequently the most immediate item on the agenda.

*8.1.2 Safety certification and verifiable constraint satisfaction*

The second obstacle is the absence of guarantees. In a model-free reinforcement learning formulation, constraints are enforced softly through reward-based penalties and early termination, which does not provide theoretical hard safety guarantees at every decision step; the remedies proposed are formal hard-constraint mechanisms including control barrier functions, primal–dual Lagrangian methods that adaptively update penalty multipliers during training, and projection-based post-processing layers that map infeasible actions onto the nearest feasible operating point [44]. A projection layer of this kind reduces constraint residuals to operationally acceptable levels but does not provide hard feasibility guarantees without a safety-net fallback; the maximum voltage violation observed was 0.0104 p.u., within the ±5% operational band but non-negligible, and formal verification is proposed to certify the surrogate's output or to identify worst-case inputs requiring solver fallback [45].

An alternative is to embed the constraint in the computational graph itself: clamping voltage magnitudes to the [0.95, 1.05] per-unit range, projecting generator outputs onto technical intervals with Sigmoid functions and validating branch flows through power-balance equations yields deterministic guarantees of constraint satisfaction, whereas omitting this module necessitates post-processing corrections that add complexity and impede compliance with real-time scheduling requirements [46]. Because power systems are safety-critical, verification algorithms have likewise been proposed to establish mathematical guarantees on the maximum error of a physics-informed equivalent relative to its differential-algebraic solution [68]. The wider tension is documented: pure learning-based methods suffer constraint violations and lack optimality guarantees, while predict-then-reconstruct schemes guarantee feasibility but may offer no computational advantage over conventional optimisation in the worst case [43]; a neural network, moreover, lacks the ability to report non-convergence and will still produce an output for an unrealisable state, remaining reliable only within the input range covered by its training data [77].

For HVAC control the binding constraint is thermal comfort, and the reported tolerance bands are not stated in a common form: an MPC deployment reported roughly 12% energy reduction with less than 0.5 °C impact on comfort [7], a deep Q-learning controller reduced energy costs by up to 45% while holding comfort within 1.2 °C of setpoints [30], and a BIM-based multi-objective twin kept predicted percentage dissatisfied below 10%, with minima of 6.2% in winter and 3.4% in summer [6]. A certifiable learned controller therefore requires the admissible envelope to be declared, and the violation metric reported, in the same terms as the guarantee being claimed.

### *8.1.3 Data requirements, benchmarking and transferability across buildings and climates*

Data scarcity is the third obstacle. Deep-learning performance relies heavily on the size and quality of the training set, and sufficient data are frequently unavailable for privacy reasons or because values are missing [56]; a physics-informed graph model required 1,346–4,042 ground-truth optimal power flow solutions generated offline, at a one-time cost of 15 min to 2.6 h, with active learning proposed to reduce the required sample count [45], while training on every possible scenario would demand an impractically large dataset [68]. Transferability is equally constrained: a cost function tuned to a particular building cannot be directly transferred to another building with a different underlying model, and output is never guaranteed to be stable outside the test set [16]. Field evidence compounds the problem, since a proprietary dataset from a single hotel in a subtropical climate limits reproducibility and generalisability [19], and because performance improvements such as HVAC energy savings of 25–50% may be over-optimistic if controllers drift, systems degrade or occupancy patterns evolve over time [74].

Benchmarking infrastructure is correspondingly thin. Existing test environments are scarce, lack options to override shading or allow window airing, typically suffice with a ten-minute sampling rate and often cover only portions of a full year, while temporal aggregation of measurements can distort the load shape and a deterministic operation scenario does not encompass the uncertainties associated with occupant behaviour [20]. The same study responds with BAC-Bench, an open test environment that lets researchers benchmark new controllers against a baseline and assess their reliability amid uncertainty in climate or building operation [20]. Two further gaps shape the agenda: comfort-oriented twin studies concentrate on university environments, so evidence from industrial and commercial settings is needed [78], and cross-climate generalisability is insufficient, particularly under mid-century extreme weather scenarios, with very few studies assessing control strategies under the heatwaves projected for coming decades [30]. Meta-learning and transfer learning are the natural methodological answers, using prior knowledge from previously trained systems so that a controller can adjust to a new operating environment with minimal additional data [59]. Rigour in model construction belongs to the same direction: of 44 reviewed chiller and heat pump studies, only 21 performed hyperparameter tuning, manual trial and error was the most common approach in eight, grid search in six, random search in four, and numerical optimisation in only three [49].

### *8.1.4 The cost of maintaining a trustworthy twin*

Trust has a recurring price. Data quality and reliability, data management and privacy, sensor limitations, and high implementation cost together with a lack of skilled personnel to maintain the system remain key hurdles to extensive deployment of digital twins in building operation and maintenance [38]. The generally substandard quality of existing models and data is a prominent obstacle, exacerbated by the substantial effort required to synthesise building systems and data-processing technologies [37], which is why calibration pipelines are explicitly designed to minimise computation cost [20].

The economics of the learned layer itself also need accounting. An approach that trains a separate model for every dynamical component achieved significant simulation speed-ups but faced barriers to adoption because users could not employ the models and tools in which they had already invested significant resources over several years [57]; a modular, plug-and-play scheme that learns separate equivalents per component and parallelises the learning task has been proposed in response [68]. Computational overhead is measurable rather than incidental: a physics-informed network required 0.66 s per epoch against 0.17 s for a vanilla network on an 11-bus system, delivering its best performance only with larger datasets [54], and the complexity of one implementation was found to be nearly quadratic, $O(n^2)$, in the number of buses, making application to million-node systems difficult and pointing to federated learning as a scaling route [42]. Incremental or continual learning that updates selected parameters with newly observed data without complete retraining would reduce downtime and computational overhead while preserving previously learned physical relationships [59], and extending an existing prediction model's knowledge through incremental learning has been named as a priority for AHU maintenance twins [76]. Because algorithm selection in such frameworks depends on the developer's previous experience and thereby influences prediction outcomes [76], and because new studies are urged to report the measurable benefits obtained and to analyse costs and savings over the twin's lifecycle [37], a comparative cost accounting of accuracy, maintenance time and compute is the appropriate evaluation frame—the reported 35% reduction in the time required to identify and resolve HVAC faults being the kind of operational quantity such an account should capture [1].

*8.1.5 Longer-term methodology: physical consistency, interpretability and continual adaptation*

The longer-term agenda concerns what learned models are made of. Three limitations persist across data-driven control of thermal storage: a lack of physical consistency in learned models, limited interpretability of black-box decisions, and insufficient cross-climate generalisability, particularly under mid-century extreme weather scenarios [30]. Fully replacing classical solvers with physics-informed networks raises concerns about model credibility and generalisation reliability, which is why hybrid-driven paradigms are increasingly favoured, with physics-informed networks assisting conventional optimisation or reinforcement learning—physics-based priors derived from power flow sensitivity introduced as a regularisation term in the actor network of a proximal policy optimisation agent reduced unsafe exploratory behaviour relative to purely data-driven reinforcement learning [79]. Achieving the balance between exploiting data-driven capability and respecting physical law is a matter of constructing the integrated loss function and tuning its hyperparameters, and adaptability to varied load conditions and renewable fluctuations still requires validation [80]. Modelling fidelity specific to the thermal plant belongs here too: latent heat behaviour is difficult to capture because conventional formulations assume a symmetric transition, whereas many phase change materials exhibit hysteresis between melting and solidification, and the choice between a PWARX model that identifies discrete thermal states and an LSTM that predicts well but lacks explicit phase-change detection is explicitly a trade-off between interpretability and predictive power [50]. The same latent-heat buffering principle is also being pursued with nanomaterial-doped PCMs outside buildings, for example in the thermal management of Li-ion batteries and its effect on the thermal-runaway mechanism [96], which widens the range of material behaviours that control-oriented models may eventually need to represent. Conceptual discipline is a precondition for progress in this direction: if the building simulation community treats any simulation model as a digital twin, the intended technological advance will be missed [4].

Autonomy and self-explanation close the agenda. With many twin models on an operation and maintenance platform and complex coupling among their parameters, analysis and prediction based on manual experience or a single artificial intelligence algorithm are difficult, so autonomous learning from real-time and historical data becomes particularly critical for adaptive model and behaviour adjustment [81]. Where large language models are proposed as the cognitive layer of such twins, the reliability of knowledge remains the primary issue, since hallucination in professional domains is unresolved, fine-tuning is prone to overfitting and consumes significant resources, retrieval-augmented generation may still produce inaccurate outputs when retrieved content is of suboptimal quality, and inference latency, data security and interpretability all constrain industrial deployment [82]. Field-validated reinforcement learning and AI/ML controllers do achieve substantial energy savings and comfort improvements, but multi-year and multi-climate testing is scarce, and the integration of predictive maintenance and human–building interfaces is often limited to single pilot studies [74]; future work is accordingly directed at combining advanced control strategies, human-centric interfaces and cybersecurity mechanisms in real-world, long-term deployments across diverse climates [74].

## 9. Conclusion

The proportional component of PID logic presumes a linear process, whereas most real processes are nonlinear, so that a single tuning maintains zero error only over a limited range: a loop tuned to perform well at high water or steam flow may be too sluggish at low flow, and one tuned at low flow may become unstable at high flow [22]. Because system gain shifts with operating conditions such as flow temperature and seasonal inlet air temperature, and because high gain tends to produce instability, there is a standing advantage in a controller that retunes itself automatically in response to changing conditions [9]. Reinforcement learning (RL) is itself an adaptive control technique that learns control policies through trial and error, receiving rewards or penalties from interaction with the environment [24]; where it has been brought to bear on the regulatory layer it has functioned as a tuner rather than a substitute, with deep RL agents achieving fast, accurate setpoint tracking and eliminating offsets on a nonlinear multiple-input, multiple-output process under time-varying uncertainty, where conventional PID failed to track the setpoint and produced large errors and offsets [3]. PID is retained within such arrangements precisely where its function still earns its place, namely where elimination of control offset conserves energy or increases comfort [8].

A framework embeds LSTM and ATCN forecasting models within a digital twin environment, directly linked to model predictive control (MPC) and reinforcement learning strategies [39].

One criterion is the accuracy–interpretability trade-off: simpler models such as linear regression and decision trees are more interpretable, complex models often achieve higher accuracy, and machine learning (ML) models are the least interpretable of the families examined, requiring model-agnostic tools such as LIME or SHAP to explain the influence of inputs on outputs [49]. Another criterion is computational cost, where ML models show considerably higher prediction times than non-ML models [49]. A further criterion is physical consistency and generalisability: learned controllers are credited with flexibility and scalability but faulted for a lack of physical consistency in learned models, limited interpretability of black-box decisions, and insufficient cross-climate generalisability, particularly under mid-century extreme weather scenarios [30]. Another consideration concerns the horizon of the decision and its verifiability—an RL agent applied to a multi-energy digital twin struggled to operate electrical and thermal storage, with limited utilisation attributed in part to the structure of a task that requires enduring a momentary penalty in charging to secure a mid-term reward in discharging [29], whereas framing the agent as a fast decision-support module whose candidate dispatch is verified by a conventional power-flow solver against operational security constraints before reaching the operator mitigates the risks of soft constraint enforcement during training [44].

A digital twin is a virtual replica of a physical asset, enabled through data and simulations, used for real-time monitoring, optimisation and decision making [10], and what distinguishes it from a simulation set-up is data assimilation: access to field measurements lets the twin use the most up-to-date load and network data, enabling significantly increased accuracy [83]. Surrogate models trained on building performance simulation output only approximate the model on which they were trained and will never be a better predictor than that model, but they allow real-time interaction, which is precisely what makes them useful for building control [4]. Physics-informed neural networks constrain the space of admissible solutions by embedding physical laws, steering training toward more interpretable predictions and improving performance in uncertain nonlinear dynamical systems [84], while the interpretability of the plant model itself bears on building user trust [49]. Architecturally, this substrate is realised in the three-layer data–model–service decomposition, whose service layer serves either as a user interface for data interaction or as part of an automated feedback loop in which modelling outcomes directly influence operation through actuators [37].

Models trained on site-specific data may require retraining for different climates or building typologies, and systems optimised for energy efficiency may not yet incorporate cost-based or carbon-aware objectives [19]. Digital twins depend heavily on accurate and complete building information modelling data, which is rarely available in operational buildings, and no single technology—IoT, ML or the twin—is sufficient on its own [28]; the same assessment finds occupant behaviour inadequately analysed despite roughly 70% of energy use occurring during occupancy, energy baseline modelling largely absent, and real-time prescriptive recommendation systems rare [28]. Reviews of digital twins for heating, ventilation, air-conditioning and refrigeration systems identify interoperability, data harmonisation and scalability as the challenges that govern deployment [23].